\documentclass[dvipsnames]{article} %
\usepackage{colm2024_conference}
\pdfoutput=1
\usepackage{booktabs}
\usepackage{graphicx}
\usepackage{enumitem}
\usepackage{wrapfig}
\usepackage{algorithm}
\usepackage{algpseudocode}
\usepackage{microtype}
\usepackage{amsmath, amssymb, amsfonts}
\usepackage{amsthm}
\usepackage{colortbl}
\usepackage[utf8]{inputenc}
\definecolor{lightgray}{rgb}{0.9,0.9,0.9}
\usepackage{caption}
\usepackage{subcaption}
\usepackage{setspace}
\usepackage{url}
\usepackage{multirow}
\usepackage{colortbl}
\usepackage{tabularx}
\usepackage{blindtext}
\usepackage{pgfplots}
\pgfplotsset{compat=1.18} 
\usepackage{tikz}
\usetikzlibrary{er,positioning,bayesnet}
\usepackage{makecell}
\usepackage{tipa}
\usepackage{siunitx}
\usepackage{nicefrac}
\usepackage{tocloft}
\usepackage{listings}
\usepackage{cleveref}
\usepackage{placeins}
\usepackage{breakcites}

\usepackage{booktabs, graphicx, xcolor, subcaption, makecell}
\usepackage{fontawesome5}

\usepackage[raster,skins]{tcolorbox} %
\usepackage{xltabular}
\usepackage{adjustbox}
\usepackage{xurl}
\usepackage{rotating}
\usepackage[normalem]{ulem}
\useunder{\uline}{\ul}{}
\usepackage{CJKutf8}
\usepackage{pifont}
\usepackage{xcolor}

\usepackage{amsmath,amsfonts,bm}

\def\eqref#1{equation~\ref{#1}}

\def\1{\bm{1}}

\DeclareMathAlphabet{\mathsfit}{\encodingdefault}{\sfdefault}{m}{sl}
\SetMathAlphabet{\mathsfit}{bold}{\encodingdefault}{\sfdefault}{bx}{n}

\newcommand*\justify{%
  \fontdimen2\font=0.4em%
  \fontdimen3\font=0.2em%
  \fontdimen4\font=0.1em%
  \fontdimen7\font=0.1em%
  \hyphenchar\font=`\-%
}

\renewcommand{\texttt}[1]{%
  \begingroup
  \ttfamily
  \begingroup\lccode`~=`/\lowercase{\endgroup\def~}{/\discretionary{}{}{}}%
  \begingroup\lccode`~=`[\lowercase{\endgroup\def~}{[\discretionary{}{}{}}%
  \begingroup\lccode`~=`.\lowercase{\endgroup\def~}{.\discretionary{}{}{}}%
  \catcode`/=\active\catcode`[=\active\catcode`.=\active
  \justify\scantokens{#1\noexpand}%
  \endgroup
}

\newcommand{\ourmodel}{\emph{ERNIE}-Image\xspace}

\title{\ourmodel{} Technical Report}

\author{
\bf ERNIE Team, Baidu \\
\vspace{0.1cm}
\texttt{ernie@baidu.com}
}

\makeatletter\def\@abstract{
We introduce ERNIE-Image, an open-source text-to-image generation model built upon an 8B single-stream DiT architecture. ERNIE-Image aims to bridge the gap between current open-source models and leading closed-source systems through more effective mining of large-scale pre-training data and improved supervision quality throughout training. During pre-training, we adopt a bottom-up data construction pipeline that combines fine-grained image categorization, rich caption annotation, aesthetic assessment, and hierarchical sampling. This strategy reduces data noise while preserving long-tail concepts and detailed real-world knowledge, providing a stronger foundation for complex generation tasks. In the post-training stage, we use a top-down data construction pipeline for high-demand scenarios, diversify prompt annotations to better match real user inputs, and apply a stabilized DPO strategy to align the model with human aesthetic preferences. We further train ERNIE-Image-Turbo for efficient 8-NFE generation and propose MT-DMD to mitigate capability drift during distillation. To make the model easier to use in practical scenarios, we equip it with a lightweight Prompt Enhancer that expands concise user intents into structured visual descriptions. In addition, we develop ERNIE-Image-Aes, an industrial-grade aesthetic model, together with ERNIE-Image-Aes-1K, a human-annotated benchmark for realistic aesthetic evaluation. Extensive qualitative and quantitative experiments show that ERNIE-Image achieves leading performance among open-source models and approaches top-tier commercial models in instruction following, text rendering, and aesthetic quality. We release the trained models and aesthetic resources to facilitate further academic research and technical progress in the AIGC community. \\

\raisebox{-2pt}{\includegraphics[height=1em]{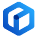}}  \textbf{Blog: } \href{https://ernie.baidu.com/blog/posts/ernie-image}{\texttt{https://ernie.baidu.com/blog/posts/ernie-image}} \\
\faGithub \hspace{0.27em}\textbf{GitHub: } \href{https://github.com/baidu/ernie-image}{\texttt{https://github.com/baidu/ernie-image}} \\
\raisebox{-2pt}{\includegraphics[height=1em]{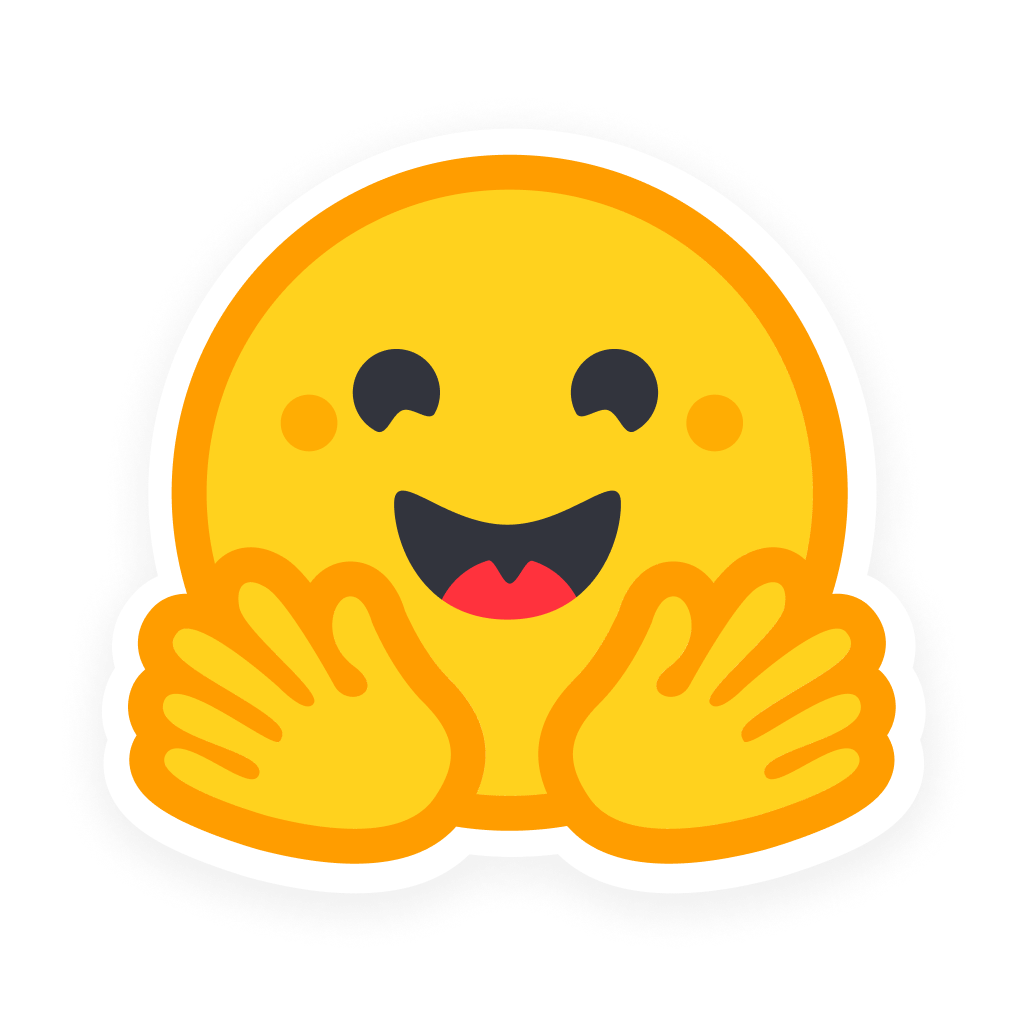}} \textbf{Huggingface: } \href{https://huggingface.co/baidu/ERNIE-Image}{\texttt{https://huggingface.co/baidu/ERNIE-Image}} \\
\raisebox{-2pt}{\includegraphics[height=1em]{latex_files/logo/hf-logo}} \textbf{Huggingface: } \href{https://huggingface.co/baidu/ERNIE-Image-Turbo}{\texttt{https://huggingface.co/baidu/ERNIE-Image-Turbo}} \\
\raisebox{-2pt}{\includegraphics[height=1em]{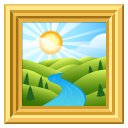}} \textbf{Image Gallery: } \href{https://ernieimageprompt.com}{\texttt{https://ernieimageprompt.com}} 

}\makeatother

\begin{document}

\maketitle

\begin{figure}[t]
    \centering
    \includegraphics[width=0.99\linewidth]{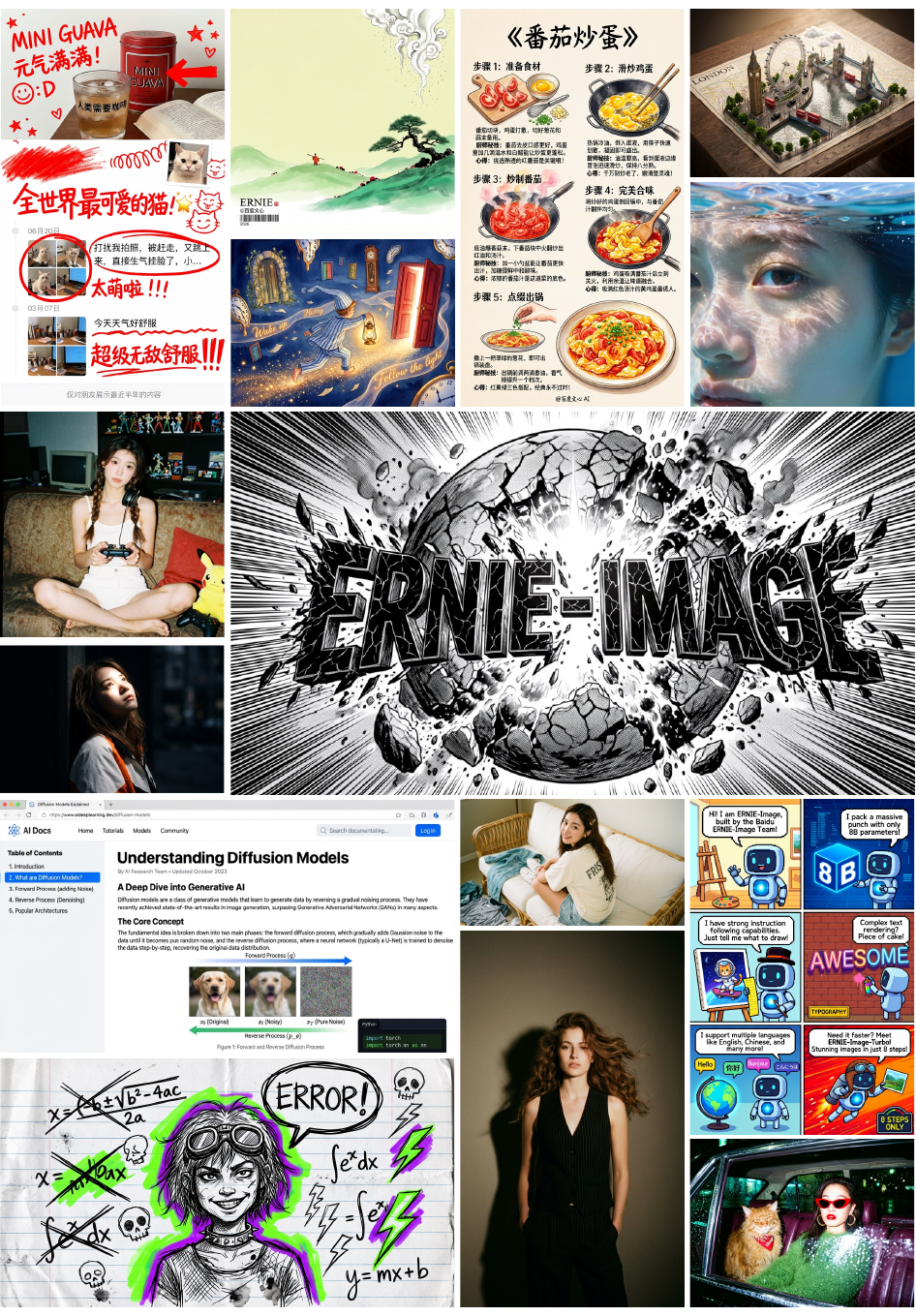}
    \caption{Showcases generated by ERNIE-Image.}
    \label{fig:examples1}
\end{figure}

\begin{figure}[t]
    \centering
    \includegraphics[width=0.99\linewidth]{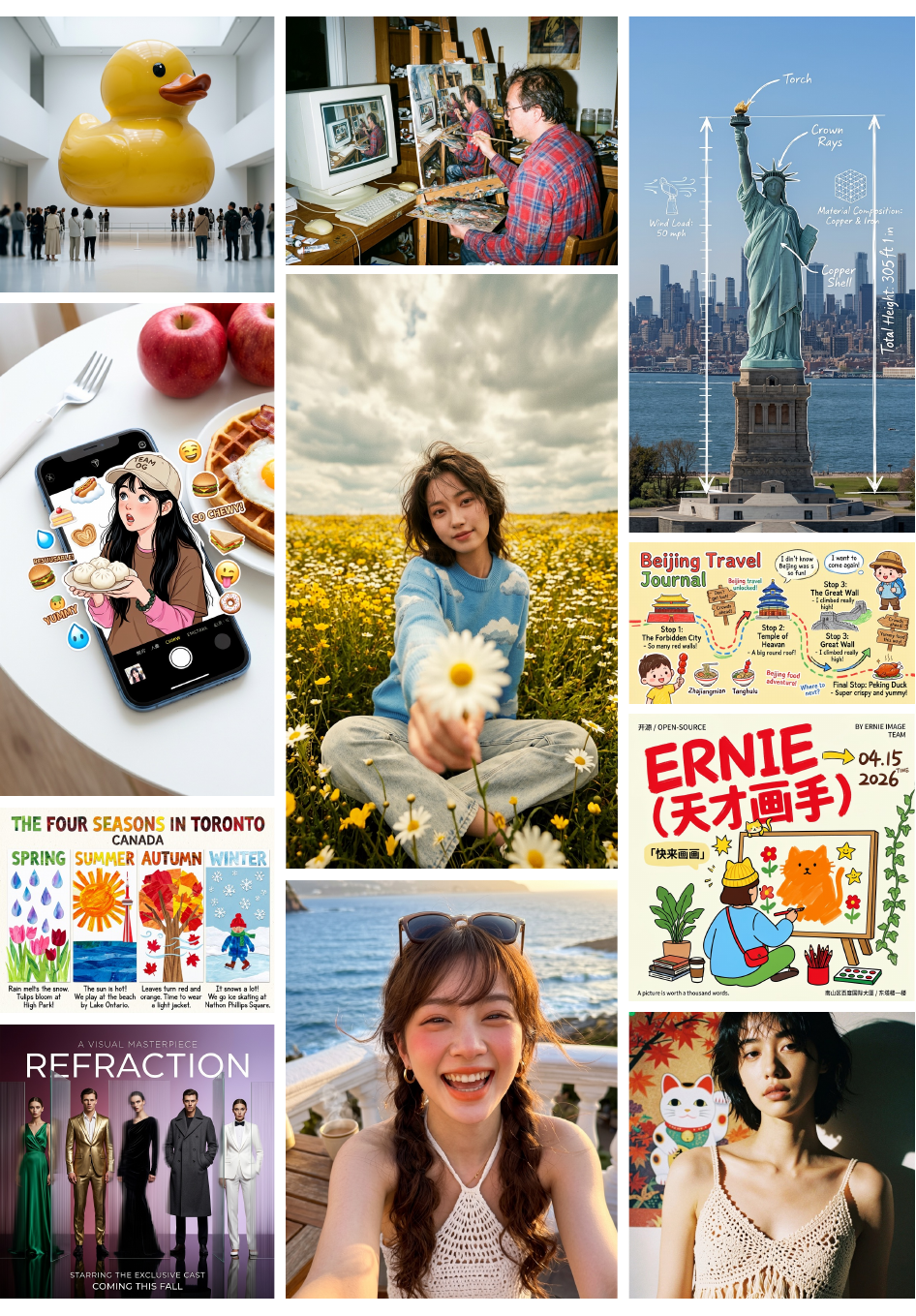}
    \caption{Showcases generated by ERNIE-Image.}
    \label{fig:examples2}
\end{figure}
\clearpage

\tableofcontents
\newpage
\section{Introduction}
In recent years, text-to-image generation has achieved remarkable milestones, evolving from generating visually plausible images to reliably synthesizing photorealistic scenes, stylized illustrations, and design-oriented content from natural language prompts. 
By transitioning from mere entertainment to practical content production, this technology has become a transformative tool of broad societal interest. 
However, as many state-of-the-art models currently adopt closed-source strategies~\citep{nanobanana2026,seedream5}, the academe and community face challenges in conducting in-depth research, private deployment, and vertical-domain fine-tuning.
Current open-source solutions struggle to effectively bridge this gap. 
While increasing parameter size—as explored by works like Qwen-Image~\citep{qwen-image} and HunyuanImage~\citep{cao2025hunyuanimage}—has driven the field forward, straightforward scaling often encounters diminishing marginal returns and introduces higher computational demands that may limit broader accessibility.
Conversely, while dense and efficient small-scale models like the 6B Z-Image~\citep{cai2025z} have demonstrated significant potential and earned strong community recognition, they still exhibit clear limitations when handling highly challenging tasks, particularly in complex instruction following and Chinese text rendering.

In this paper, we introduce~\textbf{ERNIE-Image}, which aims to realize our team’s ultimate goal: to build a model that is open-source, powerful, and easy for everyone to use. ERNIE-Image is a latent diffusion model (LDM)~\citep{rombach2022high} built upon a single-stream DiT architecture~\citep{DBLP:conf/iccv/PeeblesX23} with 8B parameters. We adopt the FLUX.2 VAE~\citep{flux-2-2025} as our variational autoencoder~\citep{kingma2013auto}, which provides a strong open-source latent space for high-fidelity image generation. Unlike other open-source models, we adopt Ministral-3(3B)~\citep{liu2026ministral} as the text encoder, further reducing the overall memory consumption and lowering the barrier to use. Through experiments, we find that the small size text encoder can still support long and complex instruction inputs while providing an effective conditioning space.

The core design objective of ERNIE-Image is to further enhance three fundamental capabilities currently lacking in open-source models: complex instruction following, text rendering, and aesthetic image generation. We observe that providing detailed image descriptions during the early training stages not only strengthens the model's complex instruction following capabilities but also significantly enriches its world knowledge. Based on this insight, rather than deliberately synthesizing domain-specific data, we adopt a more targeted annotation strategy during the pre-training data collection. For long-tail samples, we train an extremely fine-grained classification model to achieve sample balance during training. For complex instruction following, we fine-tune  a powerful VLM~\citep{qwen3} as caption model to extract structural descriptions and textual content present in the images. To elevate aesthetic preferences, we design an efficient annotation system to train a robust model \textbf{ERNIE-Image-Aes}, which was then used to clean the data. These methods substantially reduce data noise, enabling us to mine latent, fine-grained information from a massive but low-quality image pool, thereby reinforcing the model's comprehension and retention of rich real-world concepts.

In the post-training stage, we employ a top-down data construction pipeline, carefully curating highly consistent training data for high-demand scenarios. Concurrently, to further improve instruction following, we diversify the annotations of SFT data to cover various user input styles. This step enhance the model's robustness in real scenarios and mitigated the biases introduced by caption models. Sebsequently, to better align the model with human aesthetic preferences, we apply an improved DPO~\citep{DBLP:conf/nips/RafailovSMMEF23} to slightly adjust the model's distribution. We find that when both the base and reward models are sufficiently powerful, only a few DPO steps are required to reach the desired state, minimizing the reward hacking. Finally, we train~\textbf{ERNIE-Image-Turbo}, which requires only 8 Number of Function Evaluations (NFEs) to generate images. To prevent the slight capability drift that occurs when training with data subsets, we propose the MT-DMD strategy. Specifically, we utilize multiple expert variants with different capabilities to jointly provide supervision signals, ensuring the model's final optimal performance.

Another goal is to allow users to easily enjoy the model's complex instruction following capability. To this end, we equip the model with a lightweight~\textbf{Prompt Enhancer}, which expands concise user intents into richer and more structured visual descriptions before generation. This design helps ERNIE-Image better handle object attributes, spatial relationships, scene composition, textual content, and stylistic intent. As a result, the model can fulfill users' actual needs while maintaining high visual quality.

Overall, ERNIE-Image serves as both a strong open-source text-to-image model and a practical industrial solution. We summarize its key contributions as follows:

\begin{itemize}[leftmargin=20pt]
\item \textbf{Strong instruction-following and state-of-the-art text rendering.} Our model can follow diverse and complex user requirements while rendering long-form text, thereby showing practical value in real-world production scenarios such as commercial posters and advertising design. 
\item \textbf{An industrial-grade aesthetic model and high aesthetic quality.} ERNIE-Image-Aes is currently a state-of-the-art aesthetic model that can reflect aesthetic quality in a highly realistic manner. Based on this, our model can generate highly realistic photographs and simulate different films, lighting conditions, and color tones. 
\item \textbf{Open-sourced for the community.} We release all trained model weights, including ERNIE-Image, ERNIE-Image-Turbo, Prompt Enhancer, and ERNIE-Image-Aes. We also release ERIA-1K, a carefully human-annotated aesthetic benchmark, to promote open academic research and technical progress in the community.
\end{itemize}

In Section \ref{sec:performace}, extensive qualitative experiments and comparisons demonstrate the progress of ERNIE-Image. Figure \ref{fig:examples1} and Figure \ref{fig:examples2} showcase its ability to generate both realistic and design-oriented content across diverse scenarios. Furthermore, Table \ref{tab:human_preference}, Table \ref{tab:longtextbench}, Table \ref{tab:oneig_en}, and Table \ref{tab:oneig_zh} show that our model is approaching the performance of top-tier closed-source commercial models. This proves that our concise data and training strategies can break through conventional model-size limits, providing a new roadmap for future technical iterations. We publicly release model weights and aesthetic datasets to facilitate further exploration and optimization by researchers and the developer community.

\section{Model Training}
\subsection{Pre-training}

\paragraph{Data.}
In ERNIE-Image, we adopt a bottom-up data construction and sampling strategy for large-scale pre-training. Unlike top-down resampling pipelines that operate on predefined data strata, our pipeline starts from a massive raw image corpus collected from an internal data pool and progressively imposes structure through fine-grained categorization, caption enrichment, aesthetic scoring, and hierarchical sampling. This design enables broader semantic coverage while offering more precise control over category imbalance and sample quality.

We first train an image categorization model to assign each image into one of 10,000 fine-grained visual categories. This categorization step provides a scalable semantic partition of the full corpus and serves as the basis for data balancing in subsequent stages. Compared with coarse-grained sampling, such a fine-grained taxonomy allows us to better preserve long-tail concepts and prevents dominant visual categories from overwhelming the pre-training distribution.

We then use a vision-language model (VLM) to generate captions for all images. In addition to general visual description, we place particular emphasis on faithfully recognizing and describing textual content appearing inside the image. This is especially important for text-rich and knowledge-centric samples, such as slides, instructional diagrams, posters, interface screenshots, and document-style images, where missing or inaccurate text can substantially weaken image-text alignment during training.

Next, we use ERNIE-Image-Aes, our aesthetic assessment model, to assign aesthetic scores to all images. The resulting aesthetic scores are used as a unified quality signal for large-scale filtering and sampling. Rather than relying on heuristic rules alone, this learned quality estimate enables us to prioritize visually cleaner, more informative, and more visually appealing samples while preserving broad semantic diversity.

Based on the category labels and aesthetic scores, we perform hierarchical data sampling in two stages. At the inter-category level, the sampling weight of each category is determined by jointly considering its corpus size and its aggregate aesthetic quality. This design prevents high-frequency categories from dominating the training distribution, while still allocating more probability mass to categories with consistently stronger visual quality. At the intra-category level, individual images are sampled according to their aesthetic scores, so that higher-quality instances are presented more frequently during training. In combination, this hierarchical sampling strategy improves both semantic balance and sample quality, yielding a pre-training corpus better aligned with the objectives of high-fidelity and knowledge-aware image generation.

\paragraph{Training Strategies.}
We adopt a three-stage training strategy with progressively increasing resolutions. In the first stage, the model is trained at $256 \times 256$. In the second stage, training continues at $512 \times 512$. In the final stage, we further scale the training resolution to $1024 \times 1024$. At all stages, we train with diverse aspect ratios rather than restricting the model to square images only. This multi-stage, multi-aspect-ratio curriculum improves optimization efficiency at earlier stages while preserving the model's ability to generate higher-resolution images with flexible compositions and layouts.

\subsection{Post-training}
\subsubsection{Supervised Fine-Tuning (SFT)}

Following large-scale pre-training, we perform supervised fine-tuning (SFT) to further improve generation quality and instruction alignment in domains that are especially important for real-world user scenarios. In contrast to the bottom-up strategy used in pre-training, the SFT stage adopts a top-down data construction pipeline: we first identify a set of priority domains and then deliberately curate high-quality, high-consistency training data for each of them. Our current focus includes poster design, game screenshots, portrait photography, object photography, anime-style content, and other high-demand visual categories.

For each target domain, we collect data with an emphasis on both visual quality and stylistic consistency. This is important because post-training is expected not only to improve overall fidelity, but also to sharpen the model's control over composition, rendering style, subject presentation, and domain-specific conventions. Compared with broad pre-training data, the SFT corpus is therefore more selective and more tightly aligned with desired generation behaviors.

We use a vision-language model (VLM) to generate captions for all images in the SFT corpus. These captions provide structured supervision for image content, scene layout, object attributes, and visible text. High-quality captioning is particularly valuable in post-training, where the objective is not just generic image-text matching, but more precise behavioral alignment in targeted generation domains.

To better match real user inputs, we further employ a larger VLM, \emph{K2.5}~\citep{kimiteam2026kimik25visualagentic}, to rewrite and downsample the original captions into a diverse set of prompts. Specifically, for each image, we transform the caption into multiple user-facing forms with substantial variation in style, granularity, and length. These rewritten prompts include short keyword-style descriptions, natural language requests, instruction-like prompts, and more detailed compositional specifications. This prompt diversification step improves robustness to heterogeneous user expressions and reduces the mismatch between synthetic training captions and real-world prompting behavior.

\subsubsection{Direct Preference Optimization (DPO)}

To further align ERNIE-Image with human aesthetic preferences, we perform reinforcement learning via Direct Preference Optimization (DPO)~\citep{DBLP:conf/nips/RafailovSMMEF23}. By leveraging high-quality human-annotated data, DPO in our framework directly optimizes the policy by maximizing the implicit preference margin between preferred and rejected pairs.

\paragraph{Direct Preference Optimization on Flow Matching}

In the context of Flow Matching (FM)~\citep{DBLP:conf/iclr/LipmanCBNL23}, the DPO objective is reformulated from the velocity-field perspective. Inspired by~\cite{wallace2024diffusion} and ~\cite{qwen-image}, we construct DPO objective as: given a prompt hidden state $h$, a winning image $x_0^{win}$, and a losing image $x_0^{lose}$, we define the implicit reward based on the L2 reconstruction error of the predicted velocity $v_{\theta}$. The preference gaps for the policy model $\theta$ and the frozen reference model $\mathrm{ref}$ are defined as:

\begin{equation}
\begin{aligned}
\text{Diff}_{\mathrm{policy}} &= \Vert v_{\mathrm{pol}}(x_t^{win}, h, t) - v_t^{win} \Vert_2^2 - \Vert v_{\mathrm{pol}}(x_t^{lose}, h, t) - v_t^{lose} \Vert_2^2 \\
\text{Diff}_{\mathrm{ref}} &= \Vert v_{\mathrm{ref}}(x_t^{win}, h, t) - v_t^{win} \Vert_2^2 - \Vert v_{\mathrm{ref}}(x_t^{lose}, h, t) - v_t^{lose} \Vert_2^2
\end{aligned}
\end{equation}
The DPO objective is then formulated to maximize the relative gap between these margins:
\begin{equation}
\mathcal{L}_{\mathrm{DPO}} = -\mathbb{E} [ \log \sigma ( -\beta ( \text{Diff}_{\mathrm{policy}} - \text{Diff}_{\mathrm{ref}} ) ) ]
\end{equation}

where $\text{Diff}_{\mathrm{policy}}$ and $\text{Diff}_{\mathrm{ref}}$ characterize the differential $L_2$ reconstruction error between winning and losing samples for the respective models, \textbf{$\beta$} serves as a  scaling parameter, and \textbf{$\sigma(\cdot)$} is the sigmoid function.

\paragraph{Stabilizing with Anchor Losses}

A significant challenge in applying naive DPO to DiT models is \textit{reward hacking}, where the model exploits the unbounded nature of L2 loss by excessively increasing the error scale of rejected samples. This often leads to representation collapse. To mitigate this, we introduce Anchor Losses as a regularization mechanism to ground the model's fundamental generative capability:

\begin{equation}
\mathcal{L}_{\mathrm{total}} = \mathcal{L}_{\mathrm{DPO}} + \lambda_{win} \mathbb{E} [ \ell_{\mathrm{policy}}^{win} ] + \lambda_{lose} \mathbb{E} [ \ell_{\mathrm{policy}}^{lose} ]
\end{equation}

where $\ell = \Vert v_{\theta}(x_t, h, t) - v_t \Vert_2^2$, and we set $\beta=0.05$ with anchor weights $\lambda_{win}=0.35$ and $\lambda_{lose}=0.15$.

\subsubsection{Multi-Teacher Distribution Matching Distillation (MT-DMD)}

\paragraph{From DMD to DMDR}
Distribution Matching Distillation (DMD)~\citep{dmd} has emerged as a dominant paradigm for compressing iterative diffusion processes into high-efficiency, few-step generators. Modern iterations, such as \textit{Decoupled DMD} ~\citep{ddmd}, explicitly bifurcate the optimization into two orthogonal objectives to mitigate gradient interference: \textbf{C}lassifier-Free Guidance \textbf{A}ugmentation (CA), which enforces trajectory alignment with the teacher's guided path, and \textbf{D}istribution \textbf{M}atching (DM), which minimizes the divergence between generated and empirical data distributions. To stabilize this adversarial-like landscape, \textit{DMD2} ~\citep{dmd2} introduced a decoupled ratio update mechanism—executing five updates of the score model for every single generator update. 

Building upon these foundations, \textit{DMDR} ~\citep{dmdr} further enhanced the distillation process by integrating Reinforcement Learning (RL) principles and a \textbf{dynamic guidance mechanism}. To stabilize the initial optimization manifold, DMDR employs a dynamic step-aware Low-Rank Adaptation (LoRA) ~\citep{lora} scaling on the teacher score model. The real-path guidance intensity $\alpha_{real}(t)$ is modulated via a cosine schedule over a dynamic horizon $T_{dynamic}$:
\begin{equation}
\alpha_{real}(t) = \frac{\alpha_{init}}{2} \left[ 1 + \cos\left( \frac{\pi \min(t, T_{dynamic})}{T_{dynamic}} \right) \right]
\end{equation}

\paragraph{The Bottleneck of Capability Drift}
Despite the systemic stability and RL-driven refinements introduced by DMDR, these monolithic paradigms remain inherently susceptible to \textit{Capability Drift} when processing highly heterogeneous datasets. A solitary teacher, even when augmented with dynamic LoRA guidance and reward signals, often struggles to provide uniformly optimal supervision across diverse semantic domains. This fundamental limitation leads to sub-optimal convergence in specialized generative subspaces (e.g., rendering legible typography or maintaining stylized aesthetics).

\paragraph{Omni-Granular Multi-Teacher Supervision}
To neutralize these bottlenecks, we propose the \textit{Multi-Teacher Distribution Matching Distillation (MT-DMD)}. We orchestrate a committee of expert teachers $\mathcal{E} = \{E_1, E_2, \dots, E_K\}$, each implicitly optimized for distinct domain proficiencies, such as typography generation (Text-Rendering Expert), aesthetic stylization (Digital Art Expert), or macro-compositional harmony. The ensembled denoised prediction $\hat{x}_0$ is formulated via a generalized dynamic routing manifold $\mathcal{W}$:
\begin{equation}
\hat{x}_0 = \sum_{k=1}^K \mathcal{W}_k(x_t, \sigma, c, \mathcal{O}) \cdot E_k(x_t, \sigma, c)
\end{equation}
where the gating probability $\mathcal{W}_k \in [0, 1]$ adapts to the joint state space of the noisy latent $x_t$, noise scale $\sigma$, semantic condition $c$, and the specific optimization objective $\mathcal{O} \in \{CA, DM\}$.

Rather than treating sample domains and diffusion timesteps as isolated criteria, our MT routing manifold orchestrates a unified high-dimensional multiplexing strategy. When conditioned on a complex semantic prompt ($c$) demanding intricate local structures within a globally stylized context, the manifold dynamically synthesizes supervision gradients from the most congruent domain experts. This formulation enables an \textbf{asymmetric gradient topology} within a single training instance: the DM penalty can enforce global stylistic coherence by querying a Digital Art Expert, while the CA constraint independently optimizes localized spelling or anatomical fidelity via a Specialized Semantic Expert (e.g., Text-Rendering). Furthermore, the gating mechanism executes a \textbf{seamless expert handoff} along the diffusion trajectory—establishing the macro-composition via Spatial Layout experts at high-noise levels and transitioning to high-frequency rendering specialists (e.g., photorealistic lighting or material texture experts) during low-noise phases. This omni-granular paradigm effectively eliminates the capability bottlenecks of a solitary oracle, endowing the student with state-of-the-art generative versatility.

\section{Aesthetic Scoring} \label{sec:aes}
\subsection{Swiss-Tournament Based Annotation Method} \label{sec:aes_swiss}
Accurate aesthetic assessment of large-scale image datasets is a prerequisite for building high-quality training corpora for text-to-image generation. While several off-the-shelf aesthetic predictors exist, we find that they exhibit systematic biases that limit their utility in our setting. LAION-Aesthetic~\citep{laion-aes} tends to assign disproportionately high scores to AI-generated content, reflecting the distribution of its training data. ArtiMuse~\citep{cao2025uniperceptunifiedperceptuallevelimage} and UniPercept ~\citep{cao2025uniperceptunifiedperceptuallevelimage}, despite stronger general performance, suffer from poor generalization across diverse image types: for instance, we observe that ArtiMuse consistently assigns high aesthetic scores to candid photographs of everyday activities, such as amateur group dance photos taken in public parks or black\&white pictures that would not be considered aesthetically distinguished by most human evaluators. These failure modes motivate the construction of a purpose-built aesthetic scoring model tailored to our data distribution and quality objectives.

\textbf{Data Collection}. To maximize generalization and minimize distributional bias, we sample annotation candidates from our pretraining corpus with explicit category balance, spanning text-dominant images, AI-generated content, product photography, anime illustrations, professional photography, and everyday snapshots, among others. This ensures that the resulting model is not inadvertently calibrated to a limited domain.

\textbf{Annotation Methodology}. Prior work on aesthetic annotation falls into two broad paradigms: absolute rating methods, often implemented as Likert-scale scoring, in which annotators assign a numerical rating (e.g., 1–10) to each image independently, and pairwise comparison, in which annotators judge which of two images is more aesthetically pleasing. Absolute scoring is intuitive and efficient, but is susceptible to score drift, where the implicit reference point of an annotator shifts over time, causing early and late annotations to be inconsistent even within a single session. Pairwise comparison sidesteps this problem by reducing each judgment to a locally grounded binary decision, yielding more stable and reproducible labels. We therefore adopt the pairwise paradigm.

\begin{figure}
    \centering
    \includegraphics[width=0.5\linewidth]{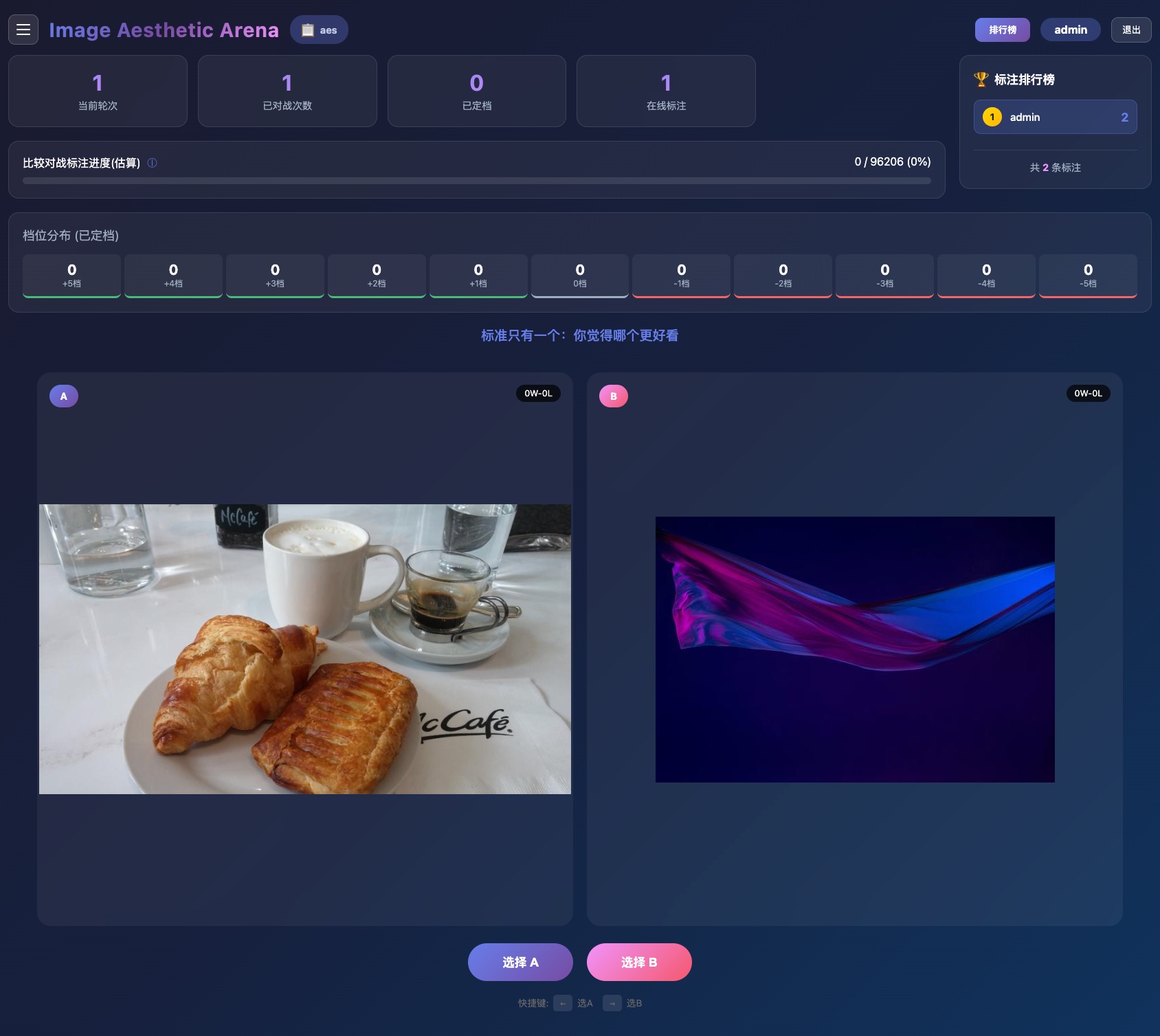}
    \caption{Screenshot from the annotation interface. Above the 2 images, there is a single line of text to remind annotators of the labeling rule for this session: "There is only one criterion: which one do you think looks more aesthetically pleasing?"}
    \label{fig:swiss}
    \vspace{-10pt}
\end{figure}

\begin{figure}[t]
    \centering
    \includegraphics[width=0.96\linewidth]{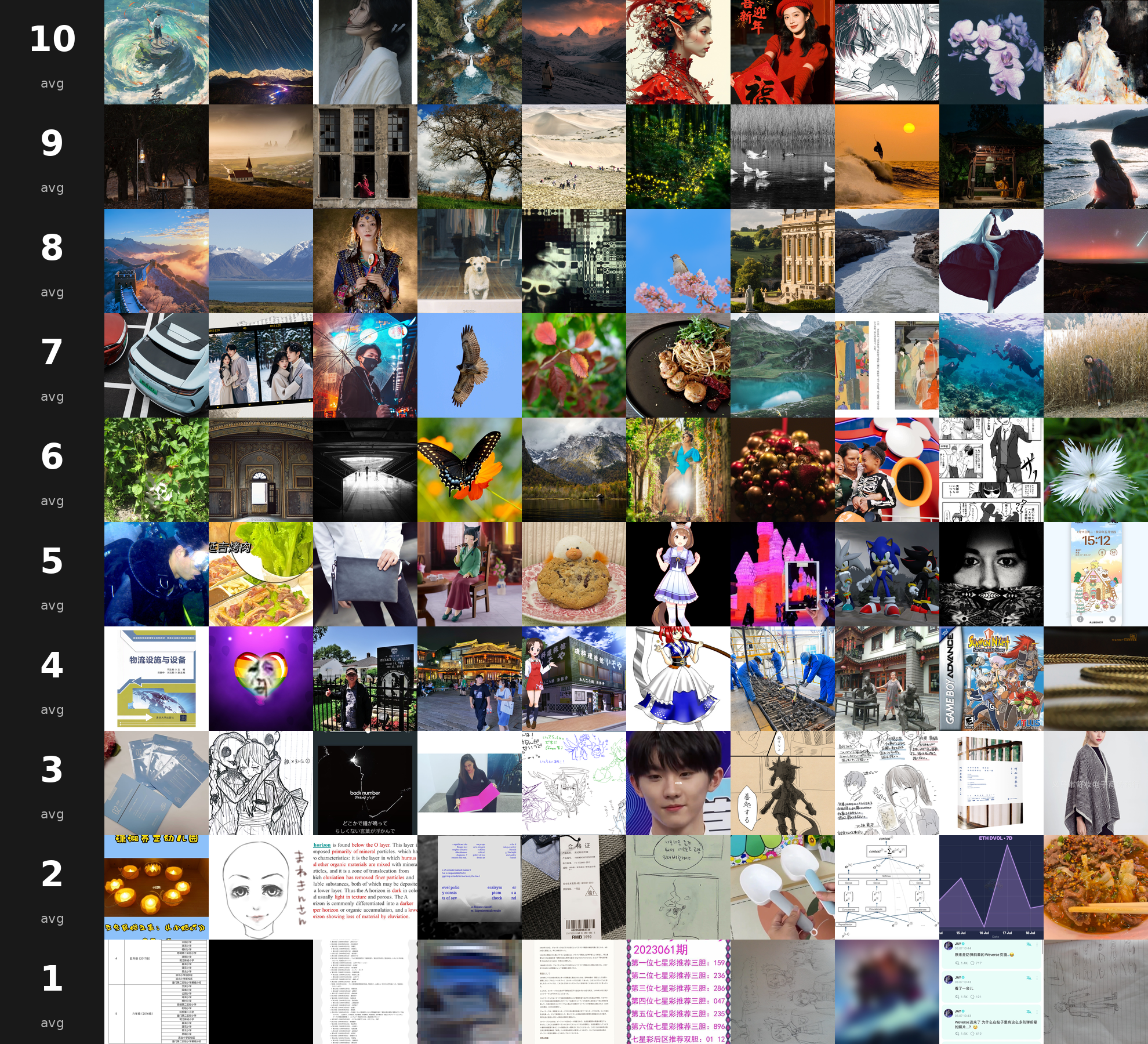}
    \caption{Preview of aesthetic annotation results.}
    \label{fig:aes_result}
    \vspace{-10pt}
\end{figure}

\textbf{Tournament Design}. Within the pairwise framework, we evaluate two candidate protocols. The first is an ELO-based rating system~\citep{elo}, in which each comparison updates the scores of both images according to the standard ELO update rule. While theoretically well-founded, we find that ELO requires a prohibitively large number of comparisons per image to produce stable rankings, making it impractical at our annotation scale. The second protocol, which we adopt, is a Swiss-system tournament~\citep{swiss}, inspired by its use in gaming (e.g., League of Legends ranked matchmaking). In the Swiss system, images are paired by current standing in each round, as shown in Figure \ref{fig:swiss}, concentrating comparisons between images of similar estimated quality and enabling reliable ranking with substantially fewer total comparisons.

\textbf{Annotation Quality Control}. During the annotation process, we identify two critical factors that substantially affect label quality. First, annotators must pass an aesthetic calibration test before participating; we find that aesthetic sensitivity varies considerably across individuals and cannot be reliably improved through training alone, making screening a necessary prerequisite. Second, we find that having multiple annotators collaboratively annotate the same pool of images within a single Swiss-system tournament produces inconsistent results, as the implicit standards of different annotators interfere with one another. Instead, each annotator independently completes a full tournament over a designated subset of images. Labels from different annotators are subsequently aggregated at the dataset level rather than merged within a single tournament.

\subsection{ERIA-1K Benchmark} \label{sec:eria}

Existing aesthetic benchmarks are predominantly constructed from curated, high-production-value image collections, often sourced from platforms such as Flickr and DPChallenge. These datasets tend to skew toward professional or semi-professional photography communities, Western photographic traditions, and visually polished content, and therefore do not fully reflect the diversity of images encountered in real-world deployment. As a result, models that achieve strong performance on such benchmarks may still exhibit systematic biases on everyday snapshots, black-and-white images, AI-generated content, or other underrepresented visual domains.

To address this gap, we introduce \textit{ERIA-1K} benchmark (\textbf{ER}NIE-\textbf{I}mage-\textbf{A}es-1K), an open-source human-annotated aesthetic benchmark designed to reflect realistic image distributions and provide a more deployment-oriented evaluation protocol.

The benchmark comprises 1,000 images spanning six categories, with proportions calibrated to approximate real-world distribution rather than overrepresenting professionally produced content. As shown in Table \ref{tab:benchmark_dist}, photography constitutes the largest portion (~49\%), reflecting its dominance in real-world image corpora. The Film Photography subset is curated for providing a high-quality film aesthetic reference. 

\begin{table}[t]
\centering
\caption{Category distribution of ERIA-1K benchmark.}
\begin{tabular}{lcc}
\toprule
\textbf{Category} & \textbf{\%} \\
\midrule
Photography  & 49.28 \\
Illustration / Anime  & 23.16 \\
Graphic Design / Posters  & 11.14 \\
Mixed Web Imagery & 10.44 \\
Film Photography & 5.42 \\
Product / Collectible & 0.56 \\
\bottomrule
\end{tabular}
\label{tab:benchmark_dist}
\vspace{-10pt}
\end{table}

All annotations are produced using the pairwise Swiss-system protocol, yielding tier labels from 1 to 10. Annotators were recruited from professional backgrounds in fine arts, design, and photography, spanning institutions including the Central Academy of Fine Arts, Sichuan Fine Arts Institute, Communication University of China, and several international programs. All annotators passed the calibration screening described in Section~\ref{sec:aes_swiss} prior to participation. This design allows ERIA-1K to serve as a high-quality held-out benchmark for evaluating the generalization ability and practical robustness of aesthetic scoring models.

\subsection{ERNIE-Image-Aes Evaluation} 
\textbf{Implementation Details} Our aesthetic scoring model is initialized from ArtiMuse~\citep{cao2025artimusefinegrainedimageaesthetics}, a 8B vision-language model, and fine-tuned on an internally annotated dataset. We build an aesthetic scoring dataset through careful human annotation, ensuring diversity along two axes. The first is the prior distribution of aesthetic quality, and the second is the distribution of image content categories, including commercial photography, film photography, everyday snapshots, anime illustrations, design graphics, and posters, among others. We find that training data diversity is critical for generalization. Models trained on narrower distributions consistently fail to generalize to out-of-distribution image categories.

\textbf{Evaluation Protocol}. We conduct our main evaluation on ERIA-1K, the held-out benchmark introduced in Section \ref{sec:eria}. Since ERIA-1K is designed to reflect realistic image distributions and is annotated by calibrated professional annotators, it provides a more deployment-oriented evaluation setting than conventional curated aesthetic benchmarks. We therefore use it as the benchmark for assessing the ability of ERNIE-Image-Aes. 

\begin{table}[h]
\centering
\caption{Comparison of aesthetic scoring models on ERIA-1K benchmark.}
\begin{tabular}{lcc}
\toprule
\textbf{Model} & \textbf{SRCC} & \textbf{PLCC} \\
\midrule
LAION AES     & 0.2944 & 0.3138 \\
ArtiMuse      & 0.4277 & 0.4704 \\
UniPercept    & 0.4533 & 0.4748 \\
\textbf{ERNIE-Image-Aes} & \textbf{0.7445} & \textbf{0.7598} \\
\bottomrule
\end{tabular}
\label{tab:aesthetic}
\end{table}

\textbf{Results}. As shown in Table~\ref{tab:aesthetic}, ERNIE-Image-Aes substantially outperforms all existing models on both Spearman's Rank Correlation Coefficient (SRCC) and Pearson's Linear Correlation Coefficient (PLCC), achieving 0.7445 and 0.7598 respectively, compared to the next best model UniPercept at 0.4553 and 0.4748.

\begin{figure*}[htbp]
  \centering

  \begin{subfigure}[t]{0.32\textwidth}
    \centering
    \includegraphics[width=0.85\textwidth]{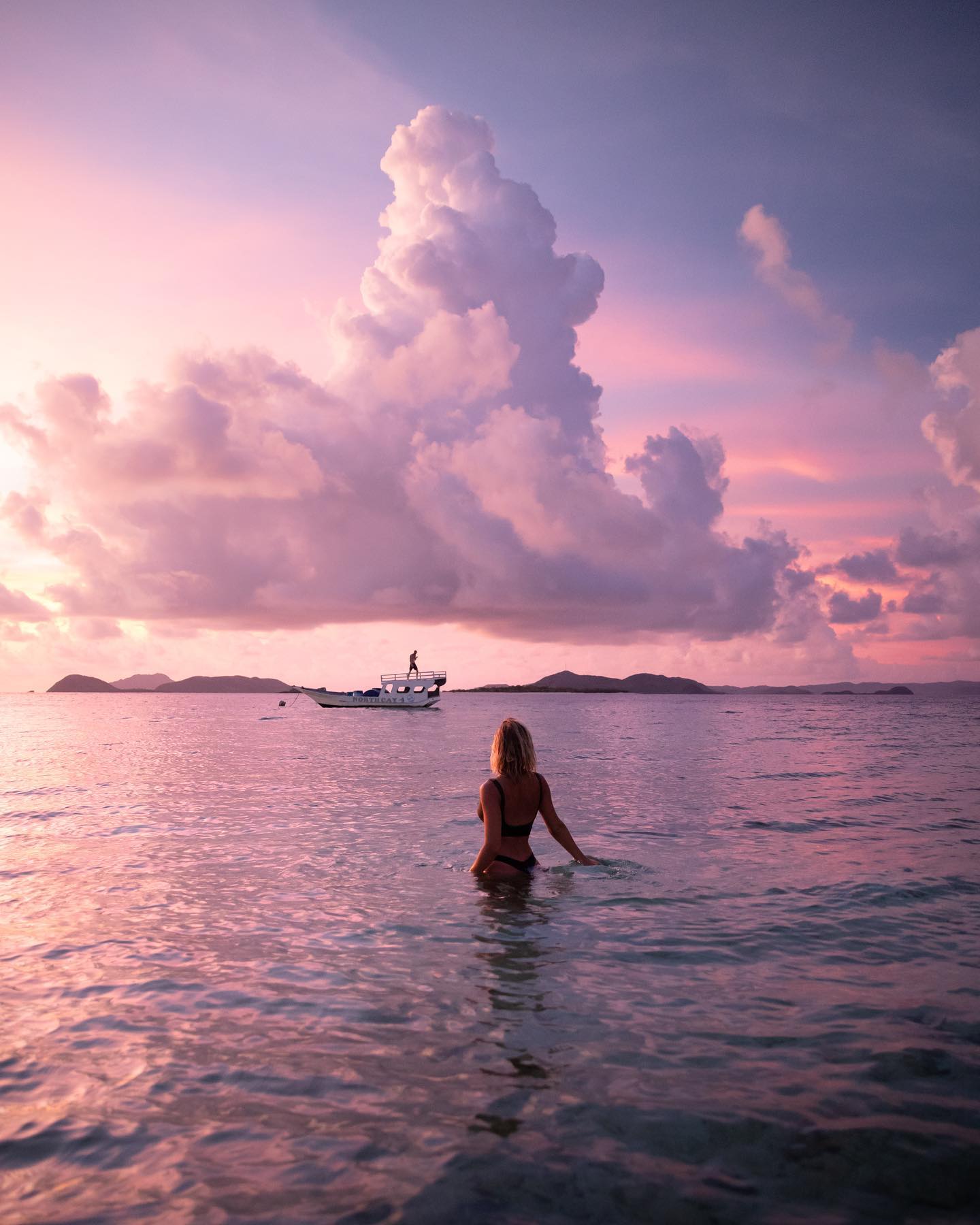}\
    \vspace{2pt}
    \includegraphics[width=0.92\textwidth]{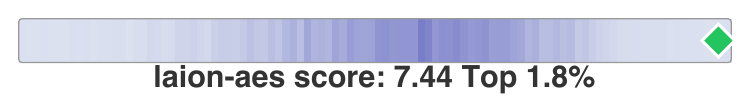}\par

    \includegraphics[width=0.92\textwidth]{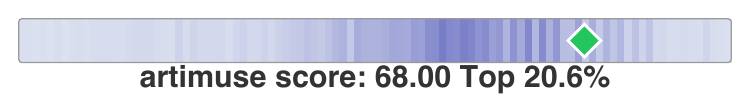}\par

    \includegraphics[width=0.92\textwidth]{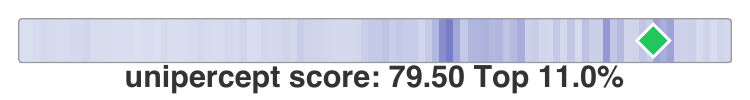}\par

    \includegraphics[width=0.92\textwidth]{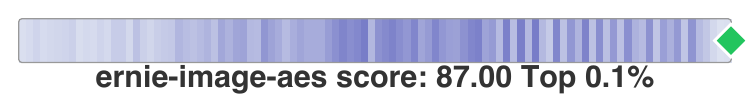}\par

  \end{subfigure}
  \begin{subfigure}[t]{0.32\textwidth}
    \centering
    \includegraphics[width=0.85\textwidth]{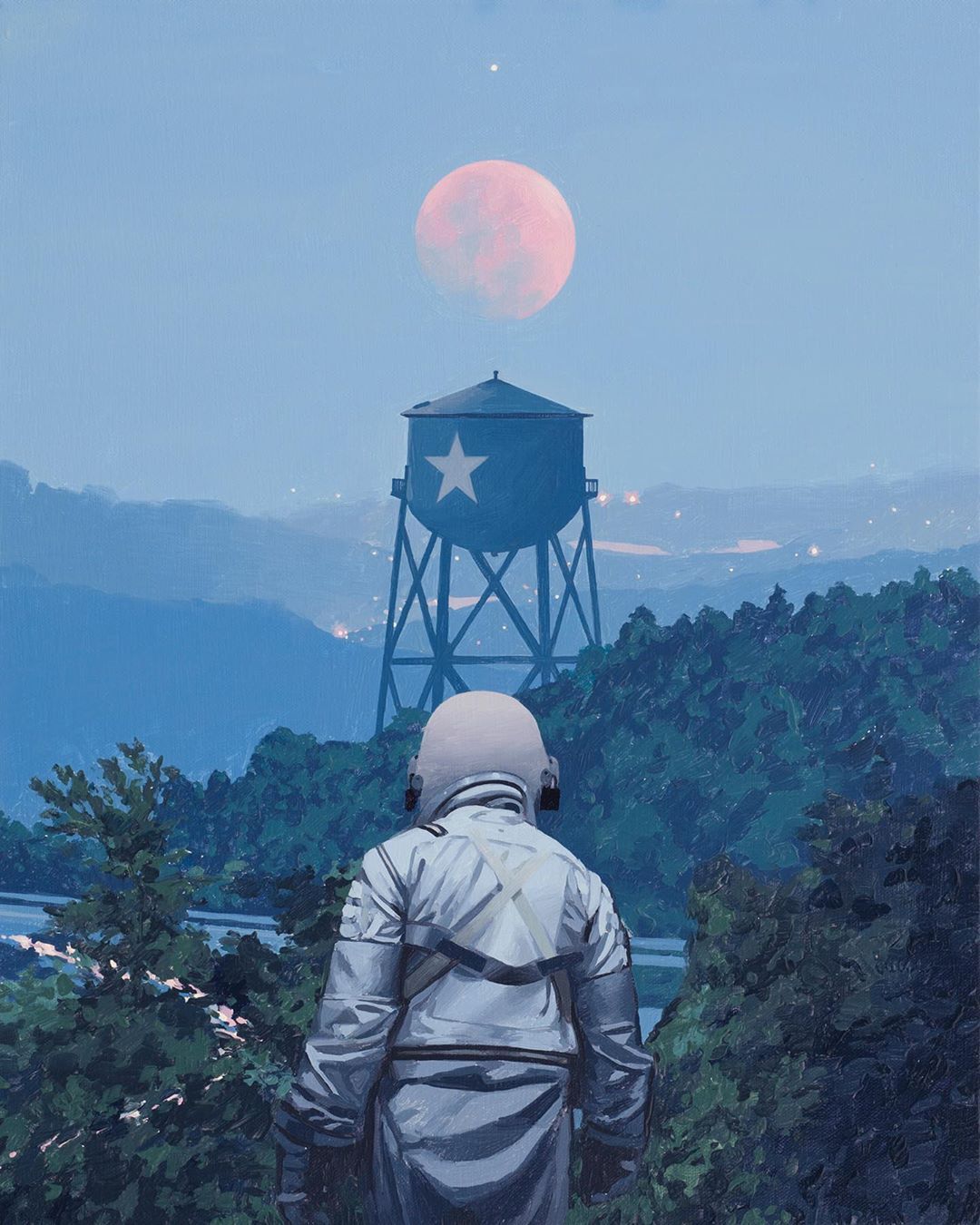}\
    \vspace{2pt}
    \includegraphics[width=0.92\textwidth]{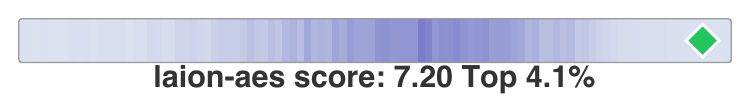}\par

    \includegraphics[width=0.92\textwidth]{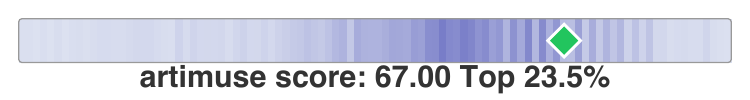}\par

    \includegraphics[width=0.92\textwidth]{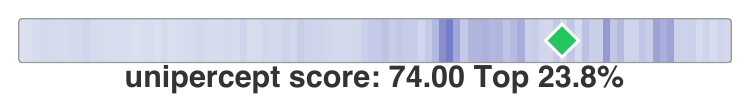}\par

    \includegraphics[width=0.92\textwidth]{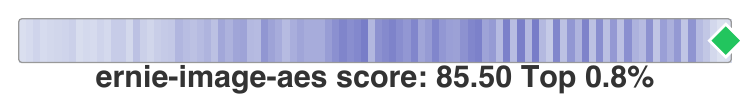}\par

  \end{subfigure}
  \begin{subfigure}[t]{0.32\textwidth}
    \centering
    \includegraphics[width=0.85\textwidth]{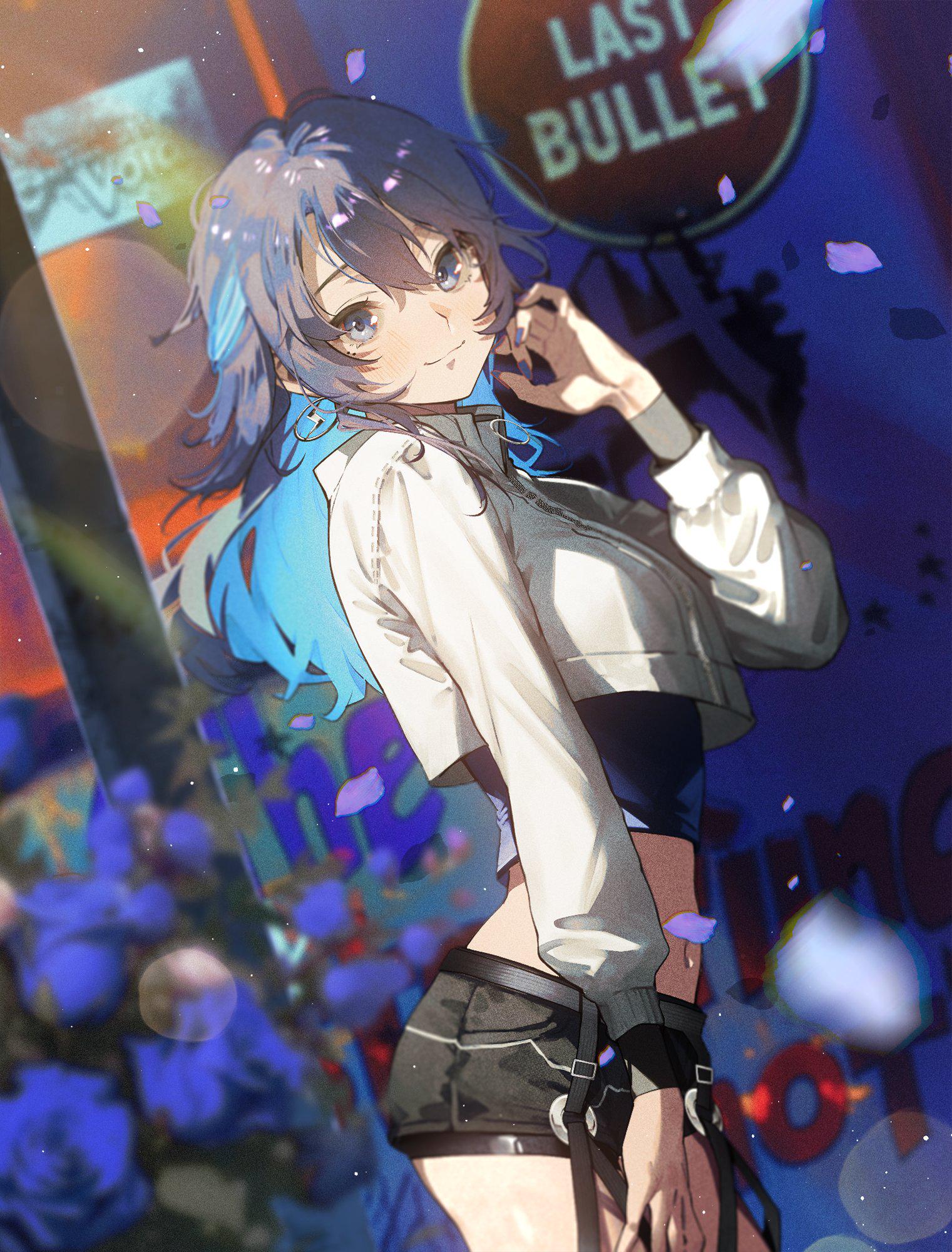}\
    \vspace{2pt}
    \includegraphics[width=0.92\textwidth]{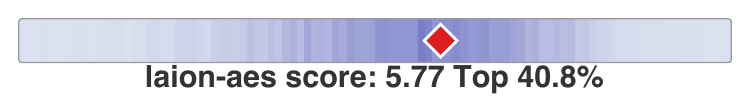}\par

    \includegraphics[width=0.92\textwidth]{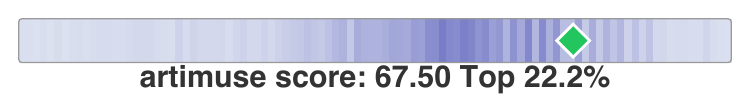}\par

    \includegraphics[width=0.92\textwidth]{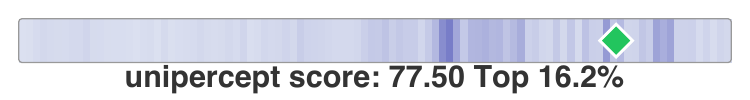}\par

    \includegraphics[width=0.92\textwidth]{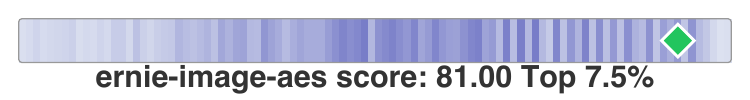}\par

  \end{subfigure}
\vspace{8pt}

  \begin{subfigure}[t]{0.32\textwidth}
    \centering
    \includegraphics[width=0.85\textwidth]{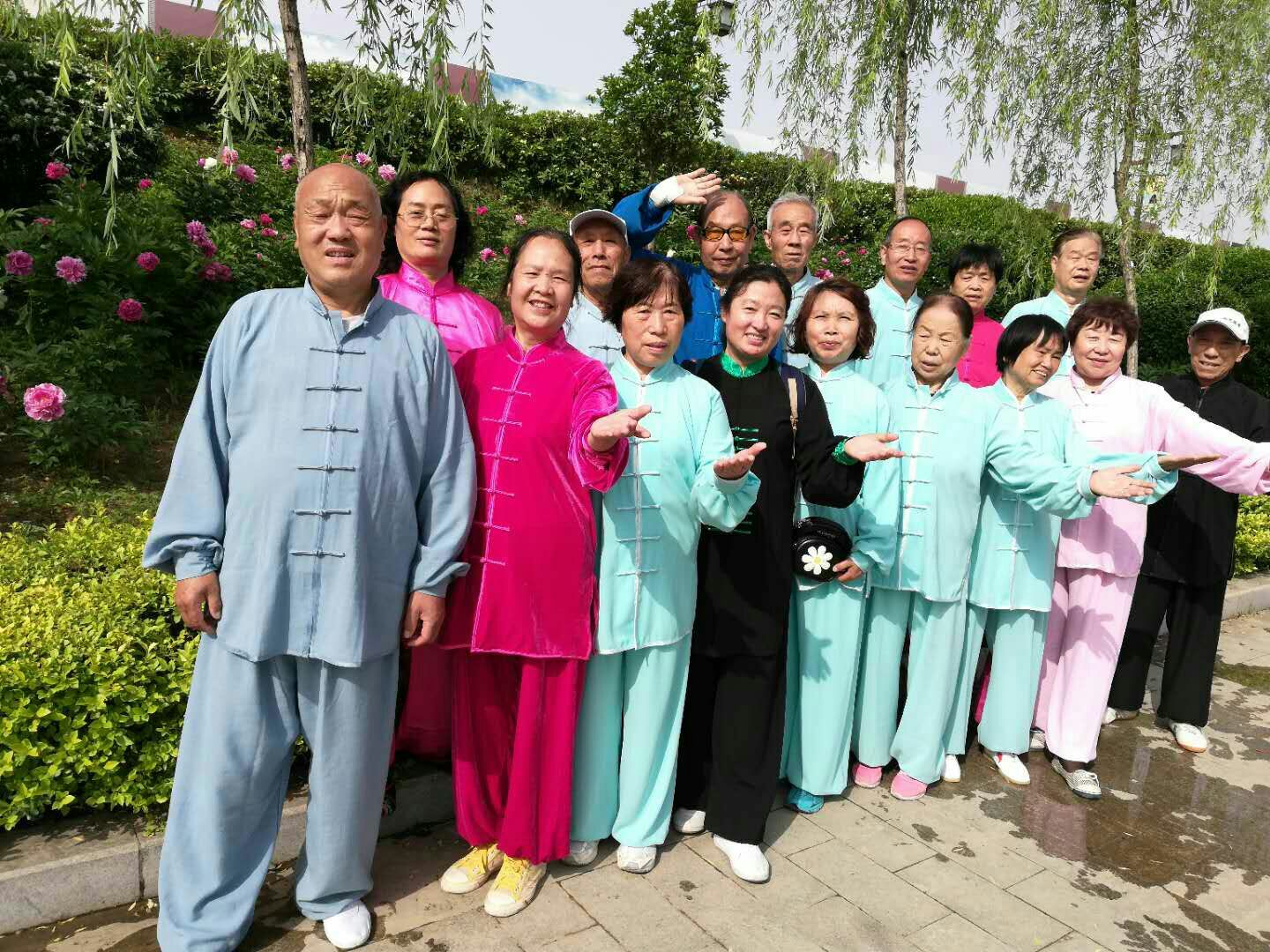}\
    \vspace{2pt}
    \includegraphics[width=0.92\textwidth]{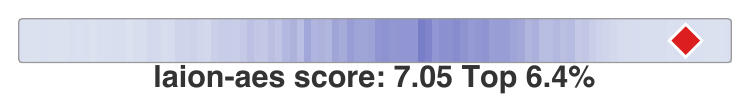}\par

    \includegraphics[width=0.92\textwidth]{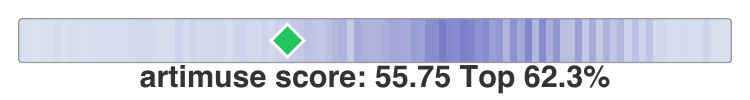}\par

    \includegraphics[width=0.92\textwidth]{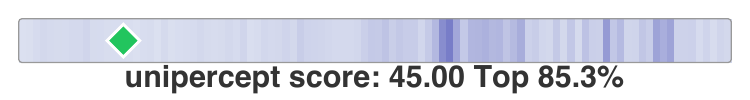}\par

    \includegraphics[width=0.92\textwidth]{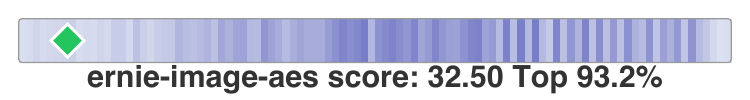}\par

  \end{subfigure}
  \begin{subfigure}[t]{0.32\textwidth}
    \centering
    \includegraphics[width=0.85\textwidth]{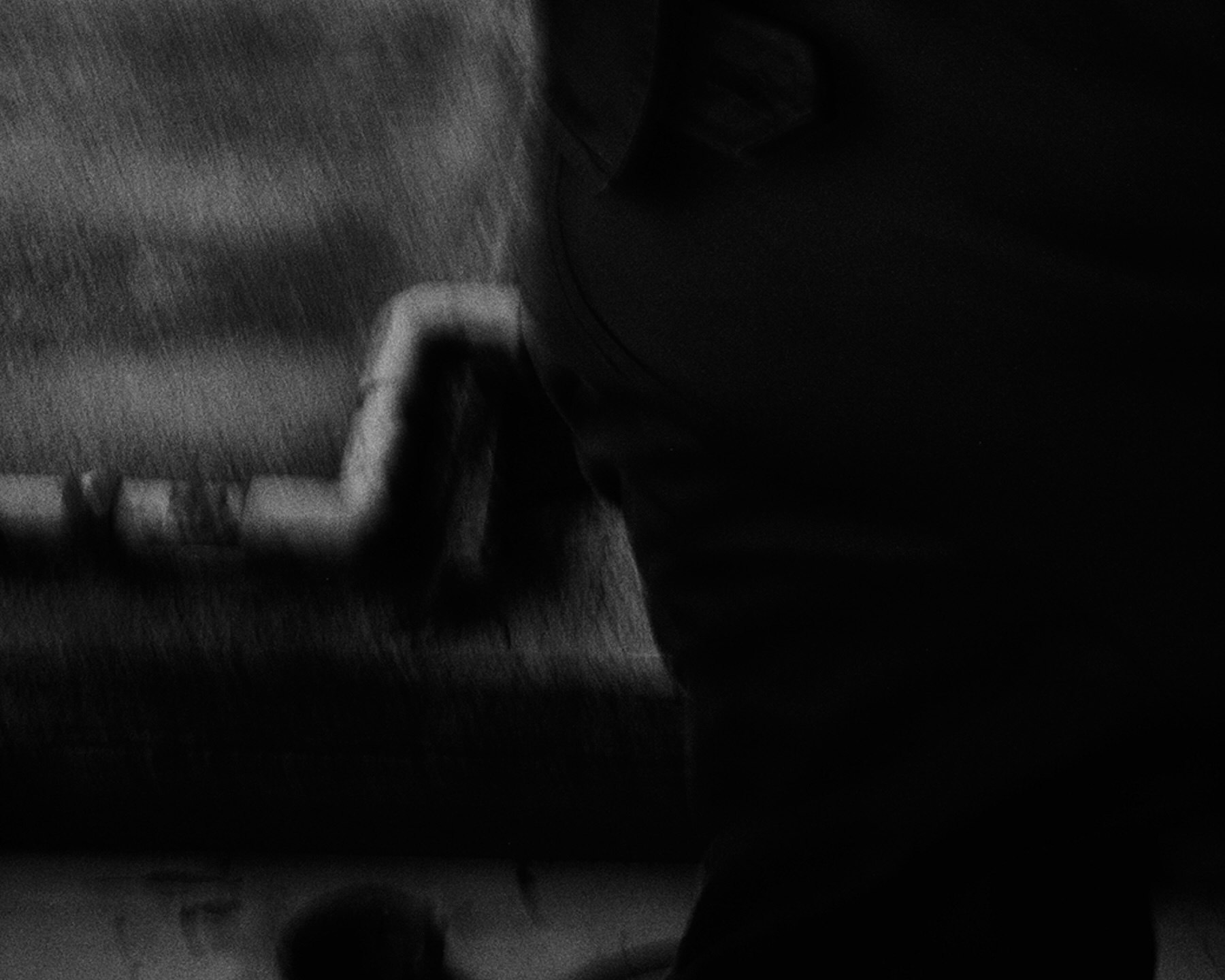}\
    \vspace{2pt}
    \includegraphics[width=0.92\textwidth]{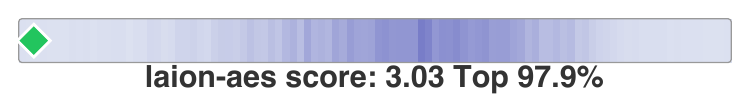}\par

    \includegraphics[width=0.92\textwidth]{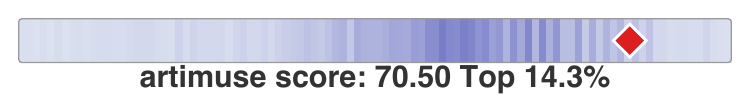}\par
    \includegraphics[width=0.92\textwidth]{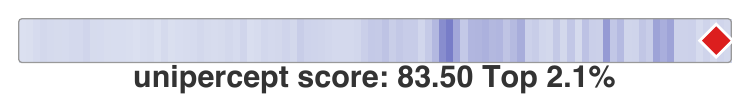}\par

    \includegraphics[width=0.92\textwidth]{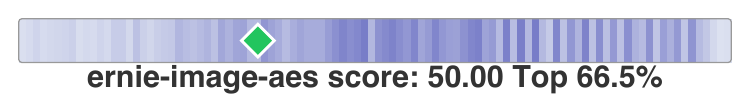}\par

  \end{subfigure}
  \begin{subfigure}[t]{0.32\textwidth}
    \centering
    \includegraphics[width=0.85\textwidth]{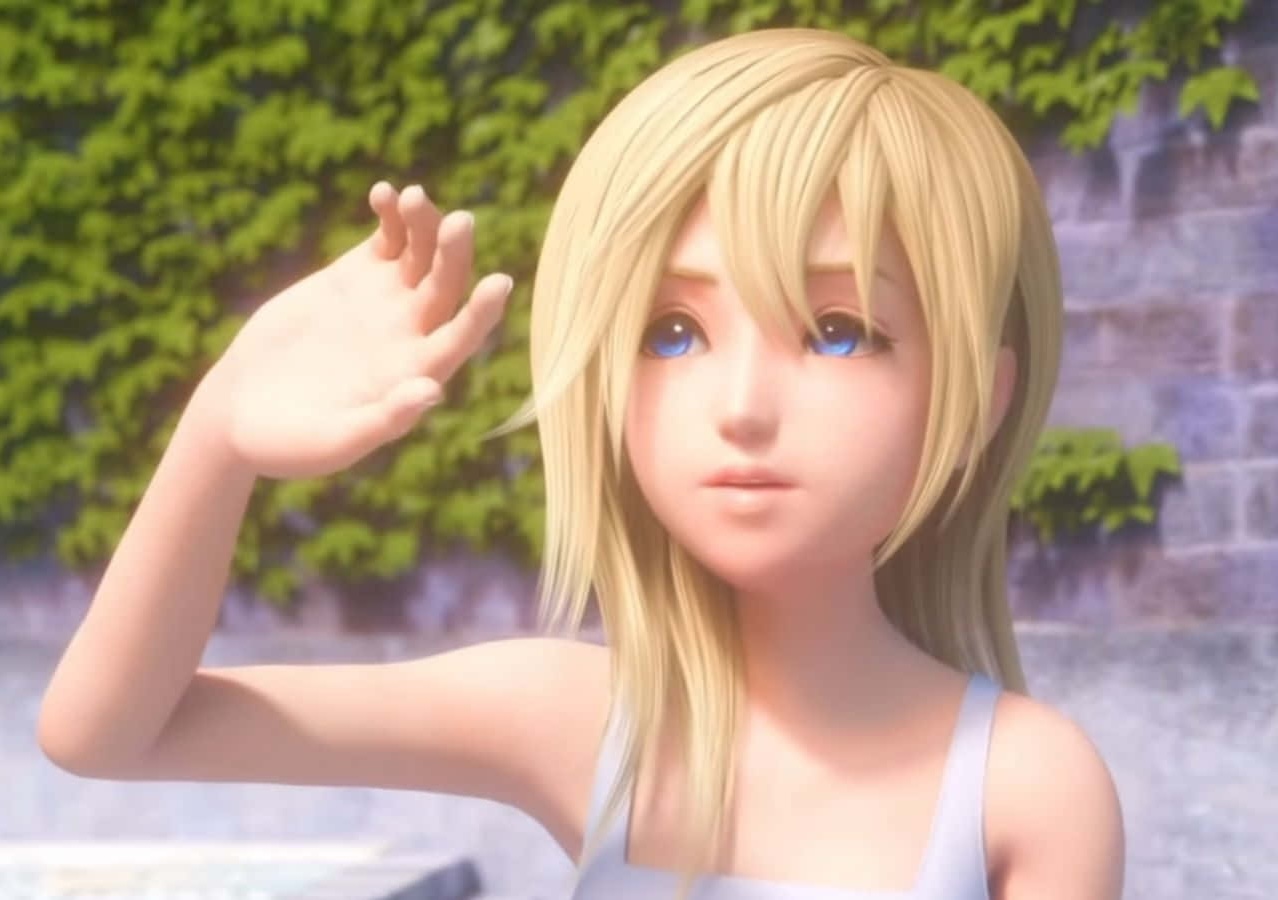}\
    \vspace{2pt}
    \includegraphics[width=0.92\textwidth]{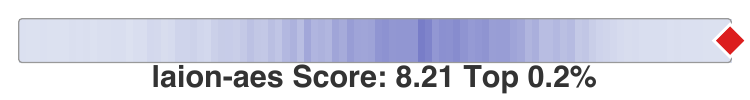}\par

    \includegraphics[width=0.92\textwidth]{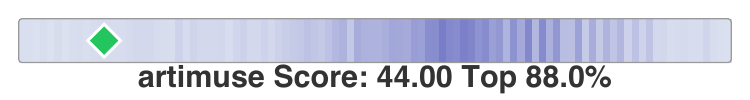}\par

    \includegraphics[width=0.92\textwidth]{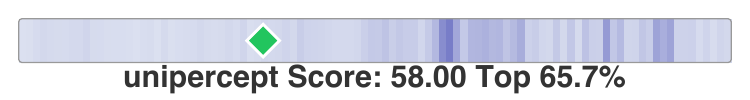}\par
    
    \includegraphics[width=0.92\textwidth]{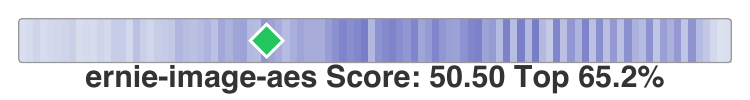}\par

  \end{subfigure}
  \vspace{8pt}
  \begin{subfigure}[t]{0.32\textwidth}
    \centering
    \includegraphics[width=0.85\textwidth]{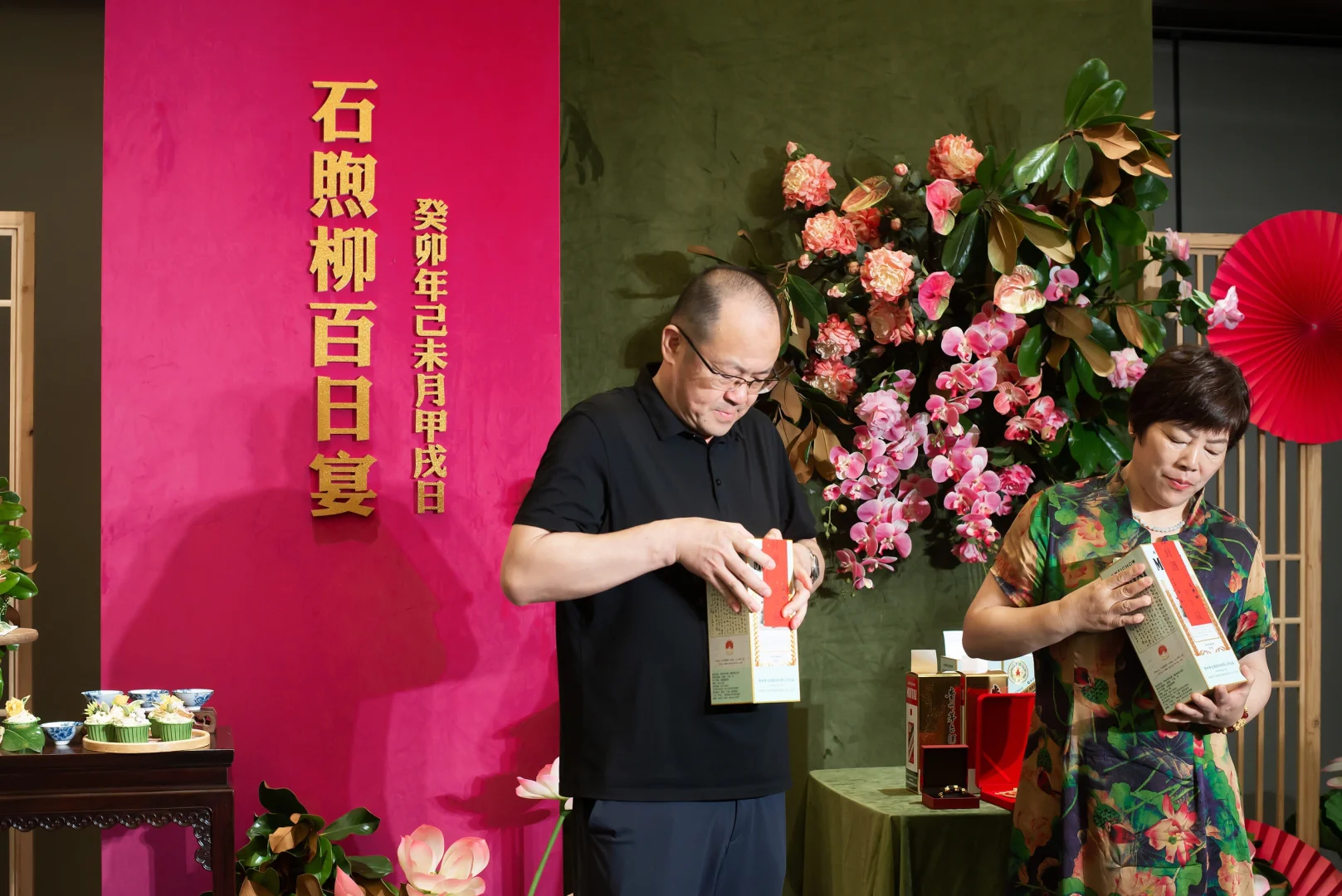}\
    \vspace{2pt}
    \includegraphics[width=0.92\textwidth]{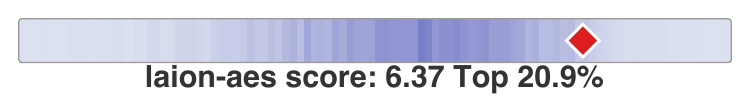}\par

    \includegraphics[width=0.92\textwidth]{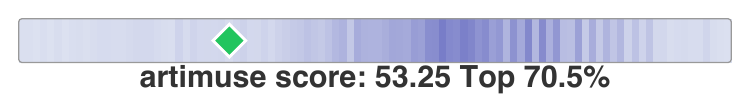}\par

    \includegraphics[width=0.92\textwidth]{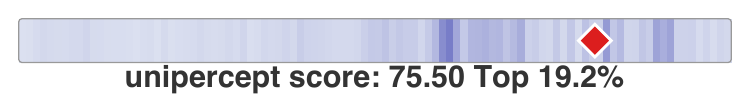}\par

    \includegraphics[width=0.92\textwidth]{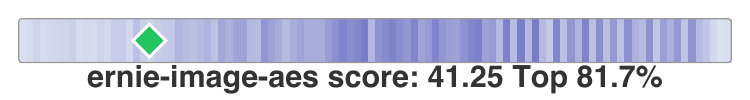}\par

  \end{subfigure}
  \begin{subfigure}[t]{0.32\textwidth}
    \centering
    \includegraphics[width=0.85\textwidth]{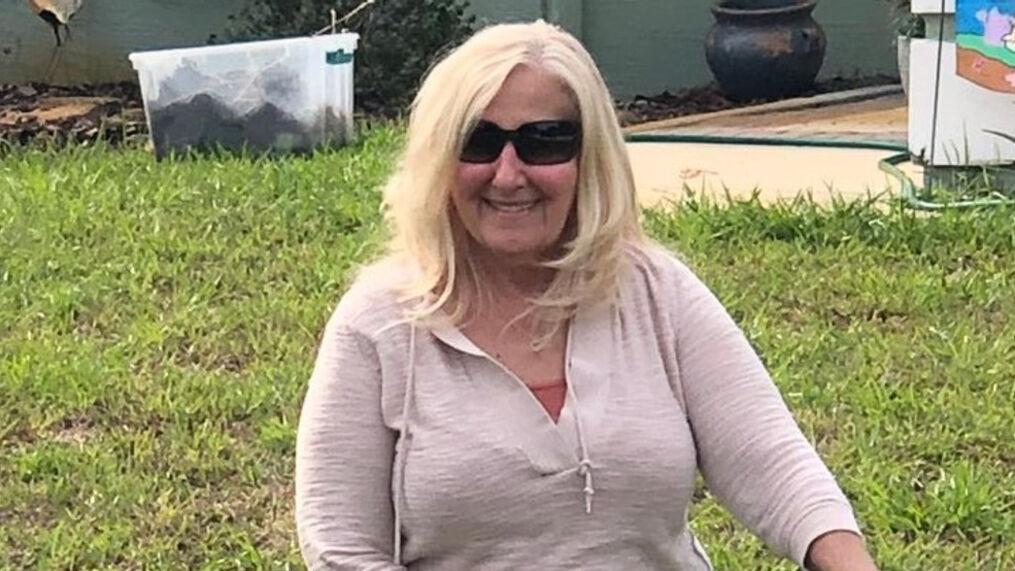}\
    \vspace{2pt}
    \includegraphics[width=0.92\textwidth]{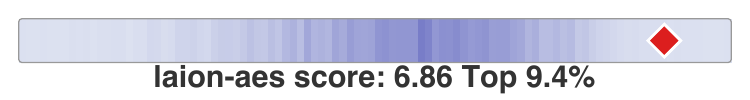}\par

    \includegraphics[width=0.92\textwidth]{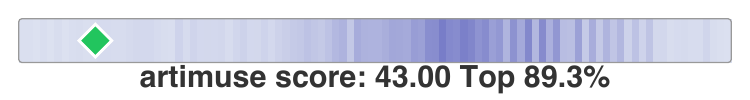}\par

    \includegraphics[width=0.92\textwidth]{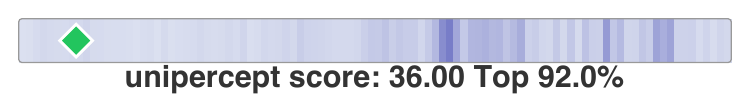}\par

    \includegraphics[width=0.92\textwidth]{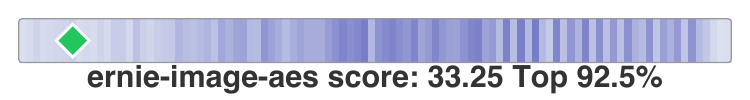}\par

  \end{subfigure}
  \begin{subfigure}[t]{0.32\textwidth}
    \centering
    \includegraphics[width=0.85\textwidth]{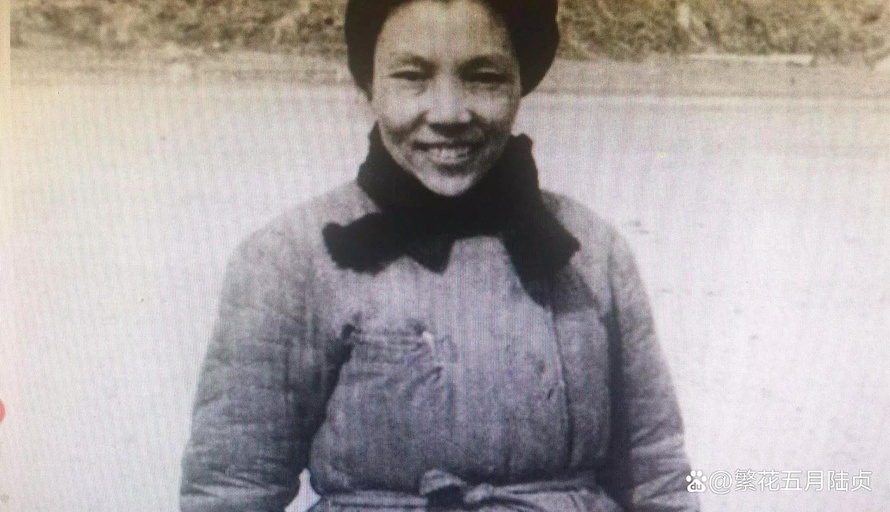}\
    \vspace{2pt}
    \includegraphics[width=0.92\textwidth]{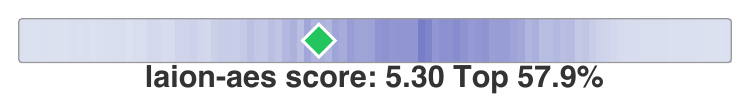}\par

    \includegraphics[width=0.92\textwidth]{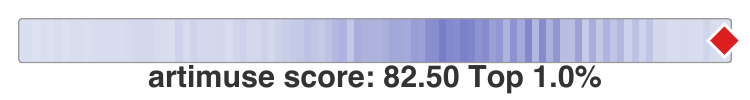}\par

    \includegraphics[width=0.92\textwidth]{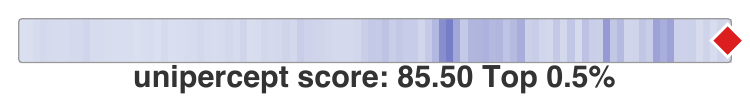}\par

    \includegraphics[width=0.92\textwidth]{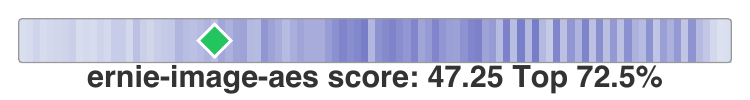}\par

  \end{subfigure}
  
  \caption{Each bar plot shows the probability density of scores predicted by the corresponding aesthetic model over our internal dataset, with lower scores on the left and higher scores on the right. The diamond marker indicates where the current image's predicted score falls within this distribution. Below each bar, we report the model name, the predicted score, and the image's percentile rank within the full dataset. For example, "Top 6.4\%" means the image scores higher than 93.6\% of all samples in terms of aesthetic quality.}
  
  \label{fig:aes-comparison}
\end{figure*}

\textbf{Bias of Off-the-Shelf Aesthetic Predictors} To expose the characteristic biases of existing aesthetic predictors, we curate a representative set of images spanning photography, casual snapshots, black-and-white images, and anime, and visualize their scores in Figure~\ref{fig:aes-comparison}. The results reveal characteristic biases for each predictor.

LAION-Aesthetic assigns disproportionately high scores to anime (image 6) and casual snapshots (images 4 and 8). Given that its training data is sourced from AI-generated content communities~\citep{laion-aes}, this bias toward non-photographic material is unsurprising.

ArtiMuse~\citep{cao2025artimusefinegrainedimageaesthetics} and UniPercept~\citep{cao2025uniperceptunifiedperceptuallevelimage} both exhibit a strong preference for black-and-white photography, with their top-scoring predictions dominated by monochrome images (images 5 and 9). UniPercept additionally over-scores casual everyday snapshots (image 7).

ERNIE-Image-Aes, by contrast, produces well-distributed predictions across image categories, assigning high scores to professional photography, high-quality anime illustrations, and film-style images while appropriately suppressing scores for low-quality snapshots. This more balanced behavior is consistent with the higher SRCC and PLCC values reported in Table~\ref{tab:aesthetic}, and reflects the deliberate diversity of our annotation pipeline.

\clearpage
\section{Performance}\label{sec:performace}
\begin{figure}[h]
    \centering
    \setlength{\tabcolsep}{0pt}
    \renewcommand{\arraystretch}{1.0}
    \begin{tabular}{@{}c@{\hspace{4pt}}c@{\hspace{4pt}}c@{}}
        \makecell[c]{\small \textbf{w/o PE}} &
        \makecell[c]{\small \textbf{w/ 3B PE}} &
        \makecell[c]{\small \textbf{w/ Larger LM PE}} \\[2pt]

        \includegraphics[width=0.312\textwidth]{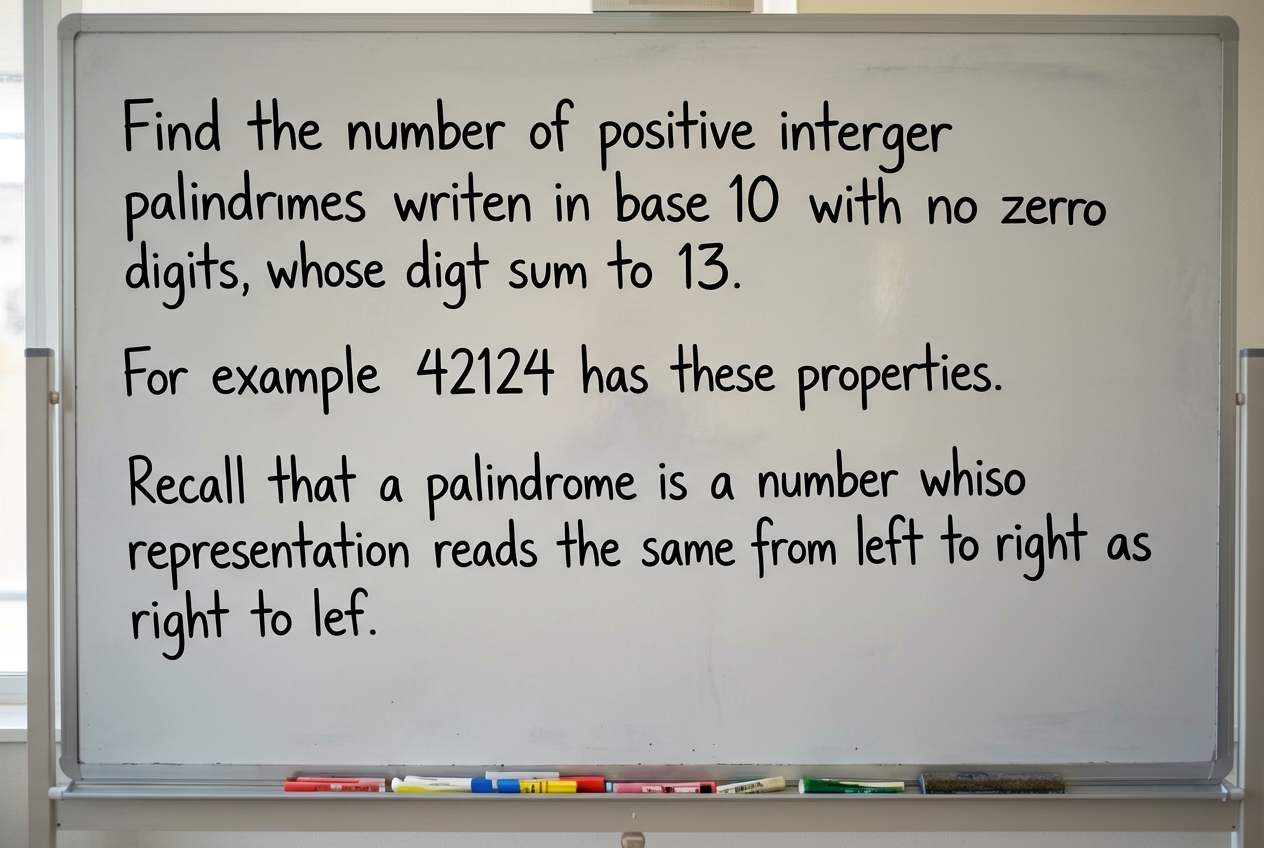} &
        \includegraphics[width=0.312\textwidth]{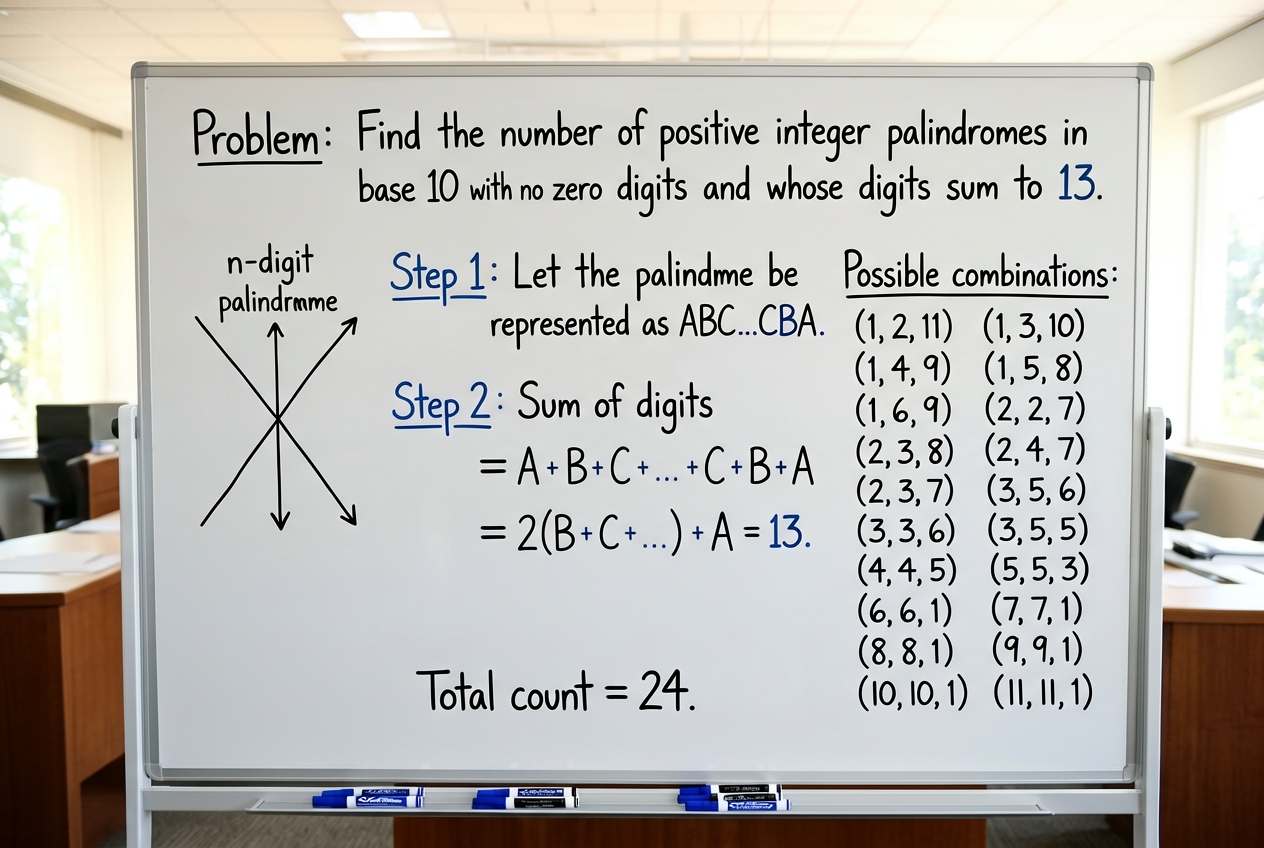} &
        \includegraphics[width=0.312\textwidth]{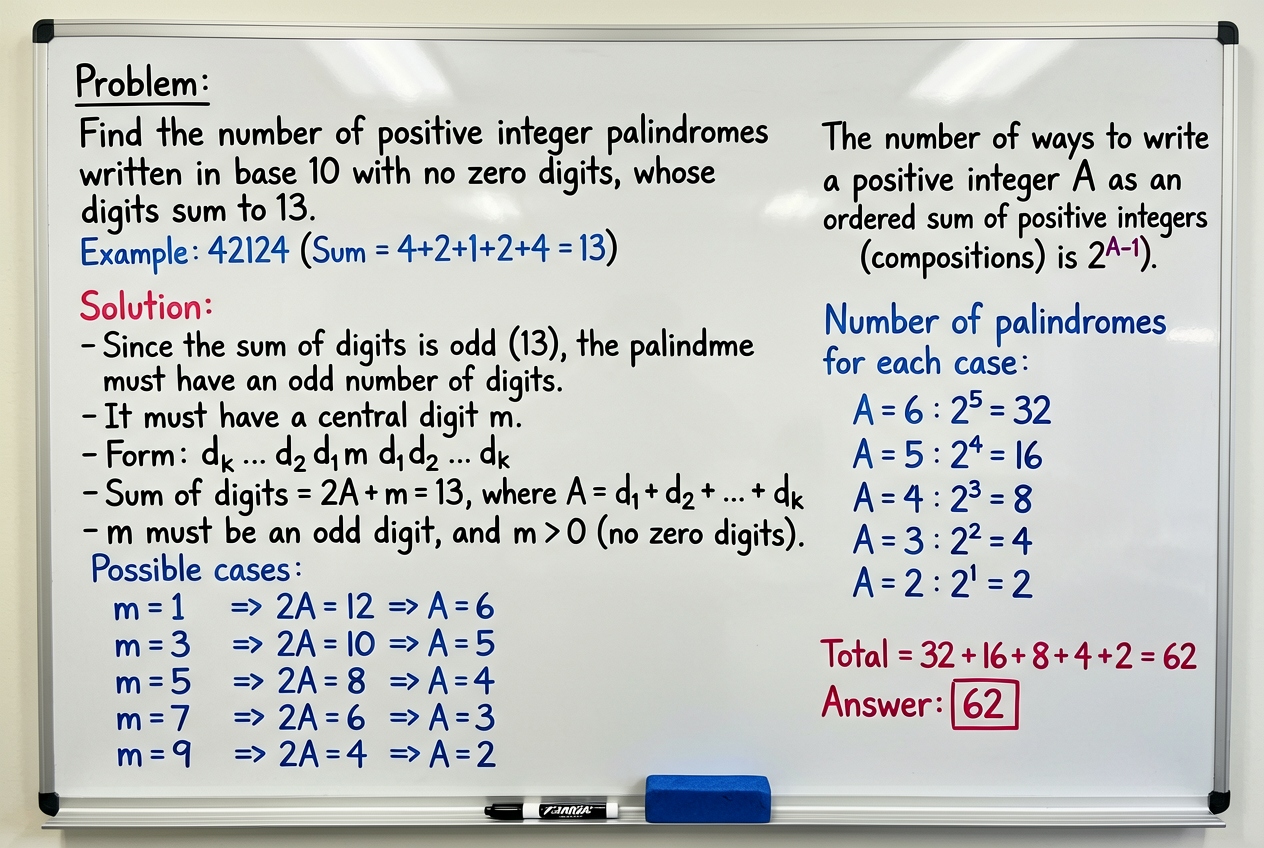} \\
        \multicolumn{3}{c}{\small \textit{``Solve this problem on the whiteboard: Find the number of positive integer palindromes written in base 10...''}} \\[4pt]

        \includegraphics[width=0.312\textwidth]{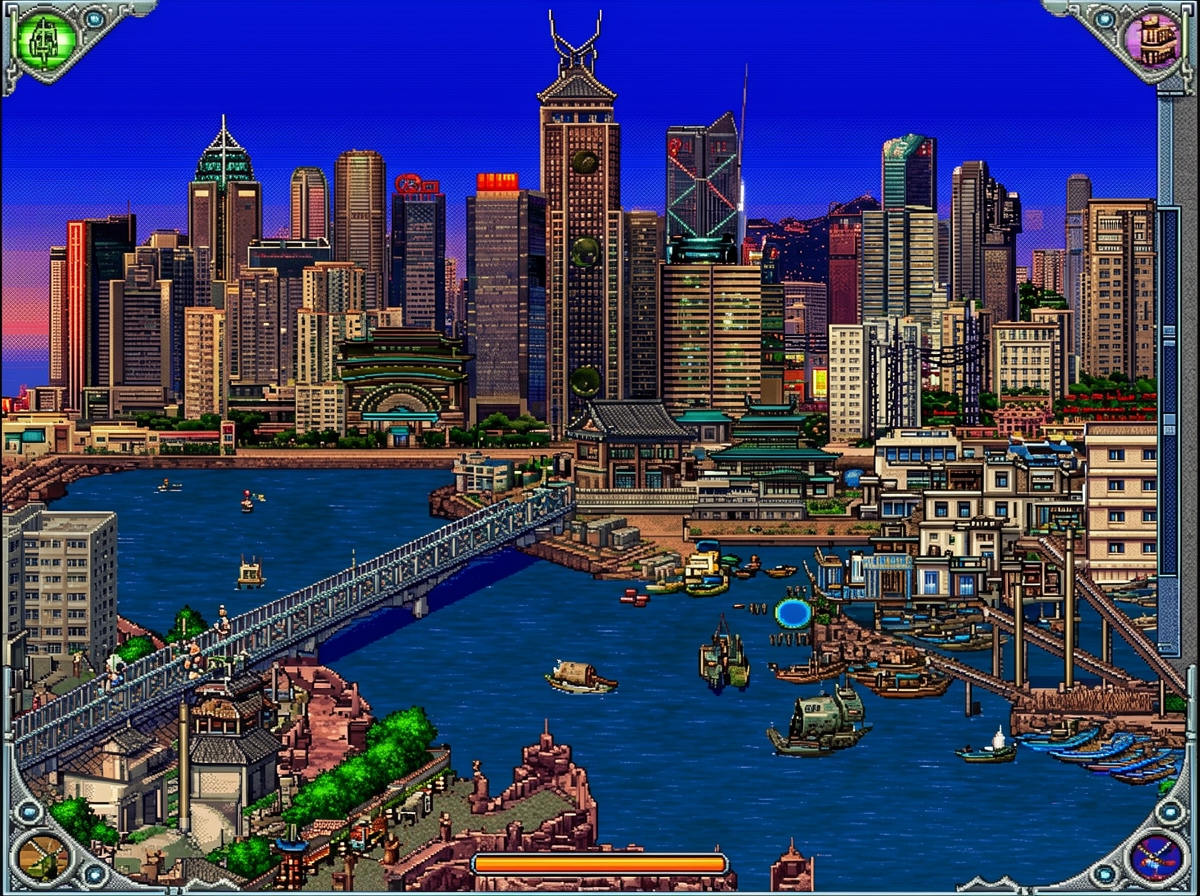} &
        \includegraphics[width=0.312\textwidth]{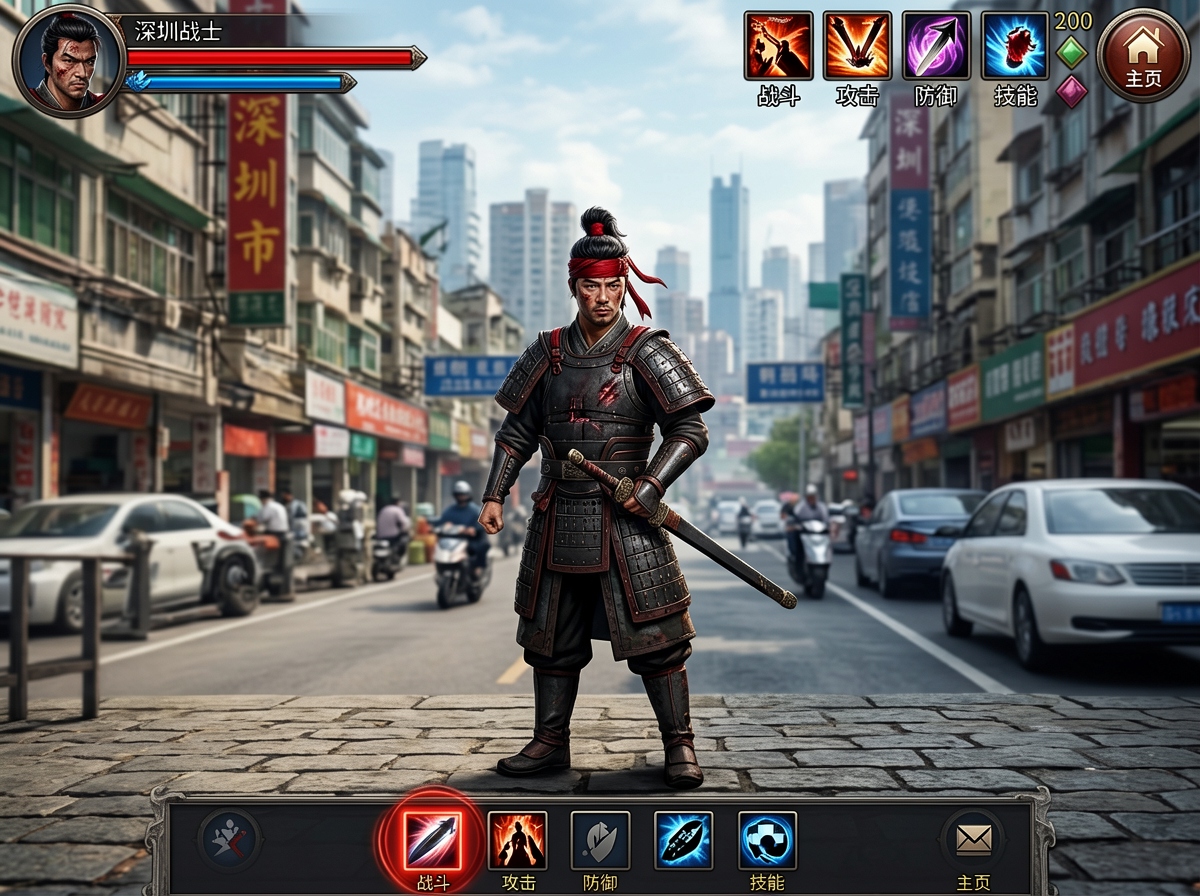} &
        \includegraphics[width=0.312\textwidth]{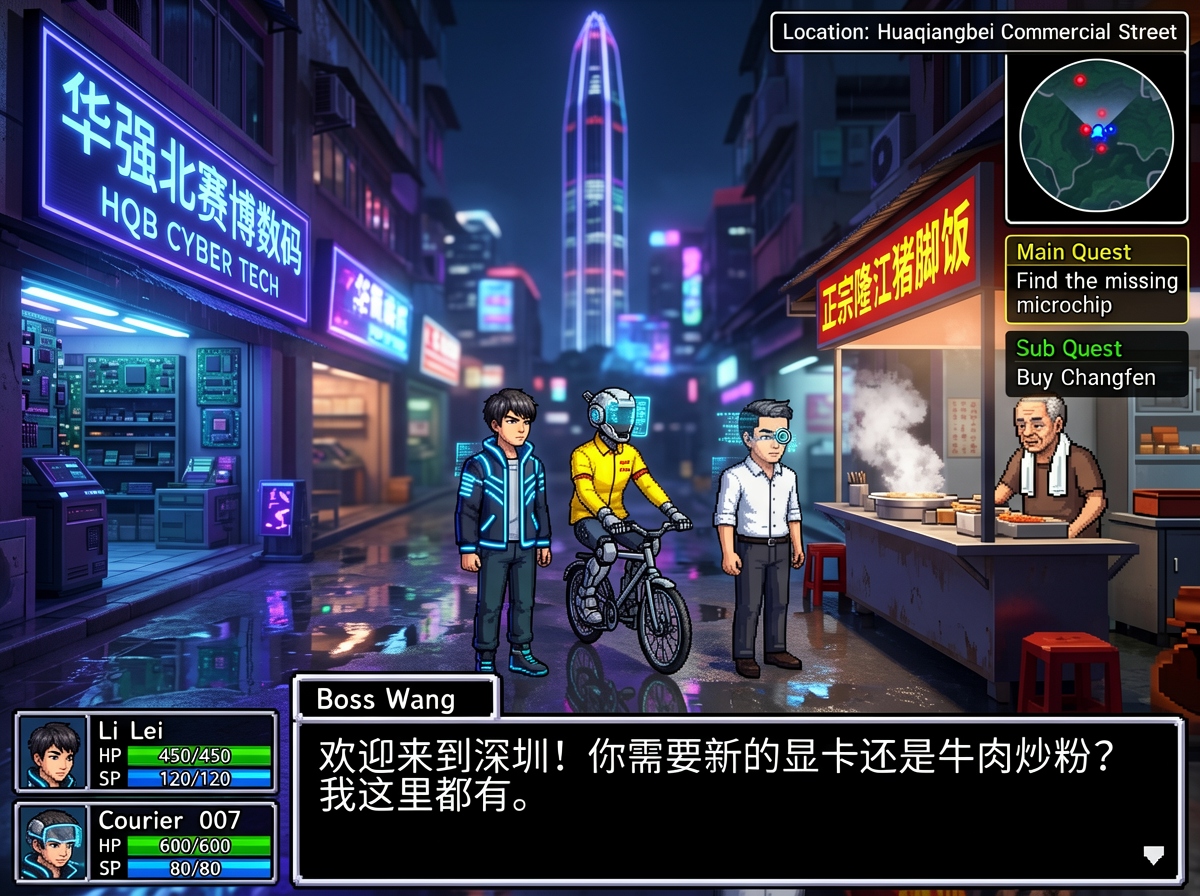} \\
        \multicolumn{3}{c}{\small \textit{``A game screenshot of an HD-2D RPG set in Shenzhen.''}} \\[4pt]

        \includegraphics[width=0.312\textwidth]{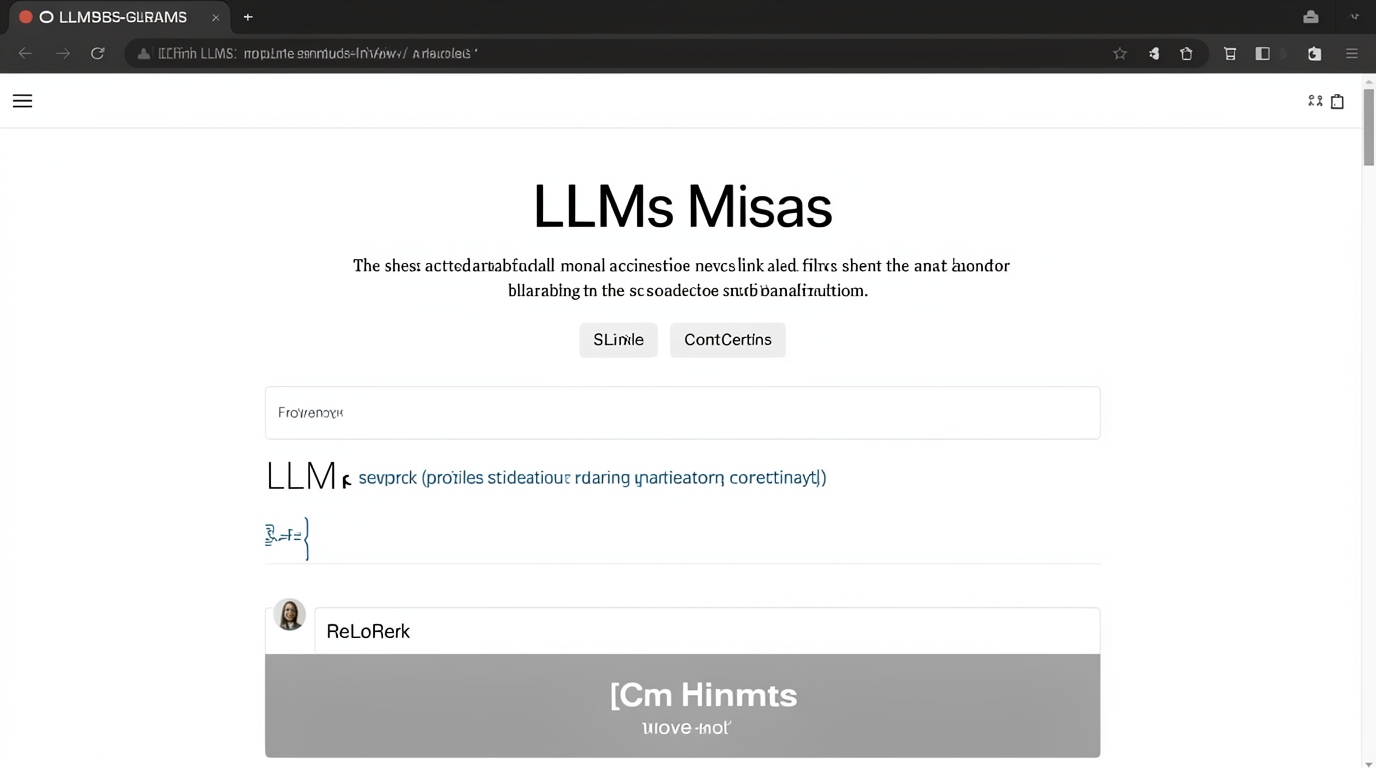} &
        \includegraphics[width=0.312\textwidth]{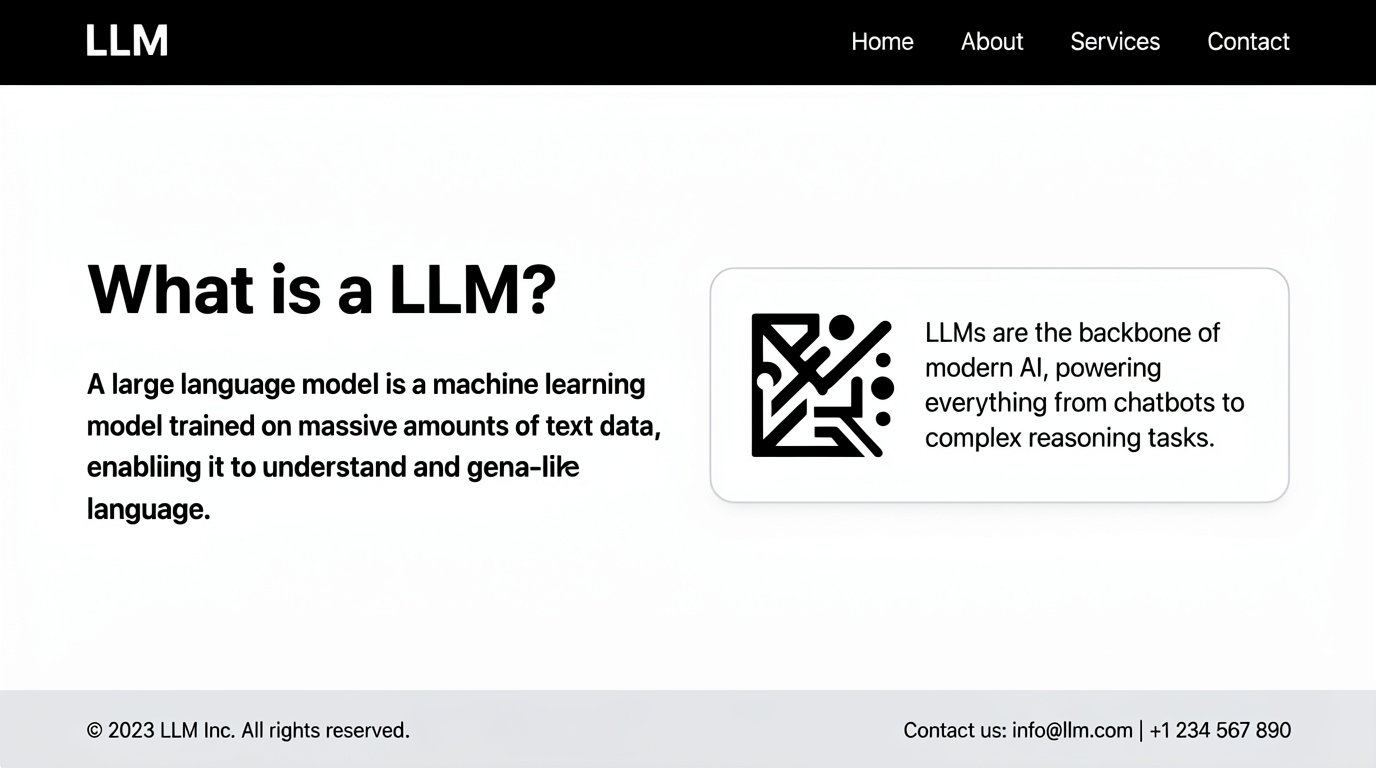} &
        \includegraphics[width=0.312\textwidth]{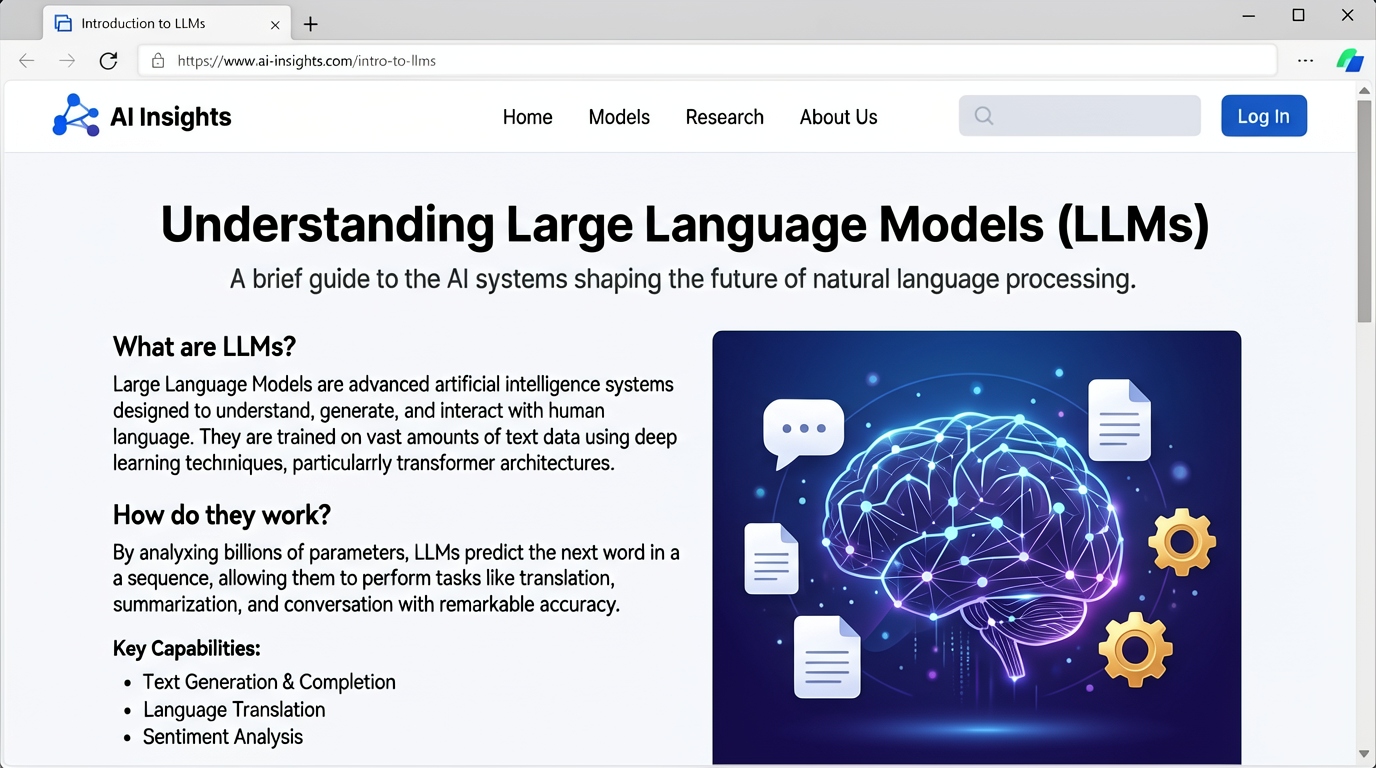} \\
        \multicolumn{3}{c}{\small \textit{``Generate a webpage screenshot that briefly introduces LLMs.''}} \\
    \end{tabular}
    \caption{Qualitative comparison of prompt enhancement on three representative tasks. From left to right: direct generation without prompt enhancement, generation with our 3B Prompt Enhancer, and generation with a Larger LM PE. Prompt enhancement consistently improves structural completeness, text fidelity, and task-specific realism, while a Larger LM PE further helps on tasks that require stronger reasoning or richer world knowledge.}
    \label{fig:pe-cases}
    \vspace{-10pt}
\end{figure}

\subsection{Prompt Enhancement Comparison}

ERNIE-Image performs best when conditioned on long, detailed, and well-structured prompts. Richer prompts usually lead to stronger instruction fidelity, more complete rendering of structured layouts, and better preservation of narrative or textual details. However, in practice, user input is often short and unspecified. To bridge this gap, we equip ERNIE-Image with a built-in lightweight Prompt Enhancer (PE), which expands concise user requests into more detailed and structured descriptions without changing the underlying user intent.

Our observations show that prompt enhancement is especially important for tasks that require the injection of latent knowledge or explicit structural planning. Figure~\ref{fig:pe-cases} illustrates this effect in three representative scenarios: solving a math problem on a whiteboard, HD-2D RPG screenshot generation, and webpage layout generation. Without prompt enhancement, the model often follows the prompt literally but only at a surface level: the requested subject is present, yet crucial structure is missing. This often appears as incomplete reasoning traces in mathematical scenes, weak game-interface semantics in RPG screenshots, or illegible and low-fidelity text in webpage-like layouts.

After applying our 3B Prompt Enhancer, the generated prompts become more explicit about layout, composition, textual content, and scene structure. As a result, the outputs are noticeably improved across all three scenarios. Whiteboard problems exhibit clearer step-by-step derivations, RPG screenshots acquire proper HUD elements and stronger genre identity, and webpage generations become substantially more legible and functional. These examples suggest that a relatively lightweight prompt rewriting model already provides a practical and reliable gain for structured image generation.

We further observe that a Larger LM PE can amplify this effect. Compared with the 3B PE, a Larger LM PE produces prompts with denser world knowledge, better reasoning, and more precise structural constraints, which is particularly beneficial for tasks that require nontrivial inference or richer domain priors. In the whiteboard example, this difference is reflected in the correctness and completeness of the mathematical derivation; in the RPG and webpage cases, it manifests as stronger environmental specificity and more realistic interface design. These results indicate that prompt enhancement is not merely a cosmetic preprocessing step, but a practical mechanism for unlocking ERNIE-Image's long-prompt generation capability.

\subsection{Human Evaluation}
\begin{table}[t]
    \centering
    \caption{Human Preference Evaluation on in-house test set. ERNIE-Image achieved the second scores overall and the first among open-source models.}
    \label{tab:human_preference}
    \resizebox{\linewidth}{!}{
        \begin{tabular}{lcccccccc}
            \toprule
            Model & Total HP & Spatial & World & Physics & Aesthetic & Style & Creativity & Knowledge \\
            & (Overall) & Structure & Understanding & Consistency & Design & Match & Ability & Accuracy \\
            \midrule
            Nano Banana 2.0~\citep{nanobanana2026}& \textbf{5.39} & \textbf{95.54} & \textbf{98.51} & \textbf{95.24} & \textbf{91.37} & \textbf{90.77} & \textbf{67.86} & \textbf{99.40} \\
            \textbf{ERNIE-Image} & \underline{5.07} & 89.88 & 94.05 & \underline{92.56} & \underline{83.04} & \underline{84.82} & \underline{62.80} & 95.24 \\
            Seedream 5.0~\citep{seedream5} & 5.03 & \underline{90.48} & \underline{97.32} & 91.96 & 80.65 & 81.55 & 61.01 & \underline{97.02} \\
            Wan2.7-Image~\citep{wan27} & 4.96 & 88.39 & 93.45 & 90.18 & 81.25 & 83.33 & 59.82 & 95.24 \\
            Qwen-Image-2512~\citep{qwen-image} & 4.78 & 85.12 & 91.37 & 88.39 & 77.98 & 79.76 & 55.65 & 95.83 \\
            \textbf{ERNIE-Image-Turbo} & 4.65 & 83.33 & 90.18 & 83.63 & 75.60 & 74.70 & 57.14 & 94.64 \\
            FLUX.2-klein-9B~\citep{flux-2-2025} & 4.59 & 81.55 & 84.23 & 84.82 & 75.00 & 79.17 & 54.46 & 85.71\\
            Z-Image-Turbo~\citep{cai2025z} & 4.56 & 81.25 & 87.20 & 79.46 & 76.19 & 78.87 & 55.36 & 89.29 \\
            Qwen-Image~\citep{qwen-image} & 4.52 & 81.85 & 87.20 & 81.55 & 72.02 & 72.32 & 57.44 & 90.48 \\
            \bottomrule
        \end{tabular}
    }
\end{table}

To evaluate the performance of ERNIE-Image in real-world scenarios, our evaluation team built an in-house test set based on user feedback. It covers various scenarios such as photography, advertising posters, and graphic design. The test focuses on core capabilities including image quality, aesthetic preference, text rendering, and instruction following. There are seven evaluation dimensions: spatial structure, world understanding, physical consistency, aesthetic design, style matching, creativity, and world knowledge accuracy. During the process, participants independently judge images from anonymous models across these dimensions, choosing ``pass" or ``fail" to calculate the final pass rates. To reflect true human preference, the process follows principles of objectivity and independence, with strict measures to prevent model leakage.

We selected several state-of-the-art open-source and closed-source models for comparison: Nano Banana 2.0~\citep{nanobanana2026}, Wan 2.7-Image~\citep{wan27}, Seedream 5.0~\citep{seedream5}, Qwen-Image series~\citep{qwen-image}, Z-Image-Turbo~\citep{cai2025z}, and Flux.2-Klein~\citep{flux-2-2025}. As shown in Table \ref{tab:human_preference}, while the closed-source Nano Banana 2.0 leads in all categories, ERNIE-Image is currently the closest open-source model. Notably, with only 8B parameters, ERNIE-Image outperforms the larger Qwen-Image series and rivals the latest commercial closed-source models. This result demonstrates the potential of small-scale models for further performance exploration and shows that our pipeline is more cost-effective than simply scaling up model size.

\subsection{Quantitative Results}

Our evaluation of ERNIE-Image focuses on two essential capabilities: general image synthesis and text rendering. To assess general generation capabilities, we conduct comprehensive evaluations on GenEval \citep{ghosh2023geneval} and OneIG-Bench \citep{chang2506oneig}. These benchmarks provide measurement of the model's ability to maintain semantic consistency and high-fidelity visual quality across complex, multi-attribute prompts. Then, to fully evaluate ERNIE-Image's capacity for complex text synthesis and cross-lingual proficiency, we conducted evaluations on LongText-Bench \citep{geng2025x}. We deliberately selected this benchmark to assess the model's ability to render dense, long-form textual content in both English and Chinese—a demanding scenario that requires precise spatial coordination and character-level accuracy. The results confirm that ERNIE-Image achieves state-of-the-art performance.

\textbf{GenEval} Table \ref{tab:geneval_final} reports the quantitative comparison of general image synthesis on GenEval \citep{ghosh2023geneval}. This benchmark evaluates models across multiple dimensions including single/two-object generation, counting, and attribute binding. As shown in the table, ERNIE-Image (w/o PE) achieves the highest overall score of 0.89, with particularly strong performance in spatial positioning and object-attribute coordination, underscoring its superior instruction-following capability.

\textbf{LongText-Bench} Table \ref{tab:longtextbench} reports the results on LongText-Bench \citep{geng2025x}, a benchmark specifically designed to assess a model’s ability to render dense, lengthy texts. Across both English and Chinese subsets, ERNIE-Image attains a competitive overall accuracy of 0.973. The remarkably narrow performance gap between the two languages illustrates ERNIE-Image’s robust cross-lingual text rendering capability and precise spatial coordination for character-level accuracy.

\textbf{OneIG-Bench} Table \ref{tab:oneig_en} and Table \ref{tab:oneig_zh} report the quantitative results on OneIG-Bench \citep{chang2506oneig}. This benchmark provides a holistic measurement of semantic alignment, reasoning, and visual quality. Across all three metrics in the English and Chinese subsets, ERNIE-Image maintains high fidelity in complex semantic reasoning and style consistency. The consistent state-of-the-art performance across these subsets confirms ERNIE-Image as a leading model in balanced image-text composition.

\begin{table}[H]
\centering
\caption{Quantitative comparison on the GenEval benchmark \citep{ghosh2023geneval}. Best results are \textbf{bolded} and second-best results are \underline{underlined}.}
\label{tab:geneval_final}
\resizebox{\textwidth}{!}{
\begin{tabular}{lcccccccc}
\toprule
Model & \makecell{Open\\Weights} & \makecell{Single\\Object} & \makecell{Two\\Object} & Counting & Colors & Position & \makecell{Attribute\\Binding} & \makecell{Overall\\Score $\uparrow$} \\
\cmidrule(lr){3-8} \cmidrule(lr){9-9}
Qwen-Image \citep{qwen-image} & \checkmark & \underline{0.99} & 0.92 & \underline{0.89} & 0.88 & 0.76 & 0.77 & \underline{0.87} \\
FLUX.2-klein-9B \citep{flux-2-2025} & \checkmark & 0.93 & \underline{0.96} & 0.83 & 0.91 & 0.72 & 0.74 & 0.85 \\
Seedream 3.0 \citep{gao2025seedream} & -- & \underline{0.99} & \underline{0.96} & \textbf{0.91} & \textbf{0.93} & 0.47 & \textbf{0.80} & 0.84 \\
Z-Image \citep{cai2025z} & \checkmark & \textbf{1.00} & 0.94 & 0.78 & \textbf{0.93} & 0.62 & 0.77 & 0.84 \\
GPT Image 1 \citep{Gpt-image-1} & -- & \underline{0.99} & 0.92 & 0.85 & \underline{0.92} & 0.75 & 0.61 & 0.84 \\
HiDream-l1-Full \citep{cai2025hidream} & \checkmark & \textbf{1.00} & \textbf{0.98} & 0.79 & 0.91 & 0.60 & 0.72 & 0.83 \\
Z-Image-Turbo \citep{cai2025z} & \checkmark & \textbf{1.00} & 0.95 & 0.77 & 0.89 & 0.65 & 0.68 & 0.82 \\
Janus-Pro-7B \citep{chen2025janus} & \checkmark & \underline{0.99} & 0.89 & 0.59 & 0.90 & 0.79 & 0.66 & 0.80 \\
SD3.5-Large \citep{SD3.5-Large} & \checkmark & 0.98 & 0.89 & 0.73 & 0.83 & 0.34 & 0.47 & 0.71 \\
FLUX.1 [Dev] \citep{flux-1} & \checkmark & 0.98 & 0.81 & 0.74 & 0.79 & 0.22 & 0.45 & 0.66 \\
JanusFlow \citep{JanusFlow} & \checkmark & 0.97 & 0.59 & 0.45 & 0.83 & 0.53 & 0.42 & 0.63 \\
SD3 Medium \citep{SD3.5-Large} & \checkmark & 0.98 & 0.74 & 0.63 & 0.67 & 0.34 & 0.36 & 0.62 \\
Emu3-Gen \citep{wang2024emu3} & \checkmark & 0.98 & 0.71 & 0.34 & 0.81 & 0.17 & 0.21 & 0.54 \\
Show-o \citep{xie2024show} & \checkmark & 0.95 & 0.52 & 0.49 & 0.82 & 0.11 & 0.28 & 0.53 \\
PixArt-$\alpha$ \citep{PixArt-a} & \checkmark & 0.98 & 0.50 & 0.44 & 0.80 & 0.08 & 0.07 & 0.48 \\
\midrule
\textbf{ERNIE-Image-Turbo (w/ PE)} & \checkmark & \underline{0.99} & 0.94 & 0.84 & 0.84 & \underline{0.80} & 0.70 & 0.85 \\
\textbf{ERNIE-Image-Turbo (w/o PE)} & \checkmark & \textbf{1.00} & \underline{0.96} & 0.79 & \underline{0.92} & \underline{0.80} & 0.73 & \underline{0.87} \\
\textbf{ERNIE-Image (w/ PE)} & \checkmark & \underline{0.99} & \underline{0.96} & 0.82 & 0.88 & \textbf{0.86} & 0.72 & \underline{0.87} \\
\textbf{ERNIE-Image (w/o PE)} & \checkmark & \textbf{1.00} & \underline{0.96} & 0.78 & \textbf{0.93} & \textbf{0.86} & \underline{0.79} & \textbf{0.89} \\
\bottomrule
\end{tabular}
}
\end{table}

\begin{table}[H]
\centering
\small
\caption{Quantitative evaluation on LongText-Bench \citep{geng2025x}.}
\label{tab:longtextbench}
\begin{tabular}{lcccc}
\toprule
Model & \makecell{Open\\Weights} & \makecell{LongText\\Bench-EN} & \makecell{LongText\\Bench-ZH} & \makecell{Overall\\Score $\uparrow$} \\
\cmidrule(lr){3-4} \cmidrule(lr){5-5}
Seedream 4.5 \citep{seedream4.5} & -- & \textbf{0.989} & \textbf{0.987} & \textbf{0.988} \\
GLM-Image \citep{glm-image} & \checkmark & 0.952 & \underline{0.979} & 0.966 \\
Nano Banana 2.0 \citep{nanobanana2026} & -- & \underline{0.981} & 0.949 & 0.965 \\
Qwen-Image-2512 \citep{qwen-image} & \checkmark & 0.956 & 0.965 & 0.960 \\
Qwen-Image \citep{qwen-image} & \checkmark & 0.943 & 0.946 & 0.945 \\
Z-Image \citep{cai2025z} & \checkmark & 0.935 & 0.936 & 0.936 \\
Seedream 4.0 \citep{seedream2025seedream} & -- & 0.921 & 0.926 & 0.924 \\
Z-Image-Turbo \citep{cai2025z} & \checkmark & 0.917 & 0.926 & 0.922 \\
Seedream 3.0 \citep{gao2025seedream} & -- & 0.896 & 0.878 & 0.887 \\
GPT Image 1 [High] \citep{Gpt-image-1} & -- & 0.956 & 0.619 & 0.788 \\
FLUX.2-klein-9B \citep{flux-2-2025} & \checkmark & 0.864 & 0.218 & 0.541 \\
BAGEL \citep{bagel} & \checkmark & 0.373 & 0.310 & 0.342 \\
Kolors 2.0 \citep{Kolors2.0} & \checkmark & 0.258 & 0.329 & 0.294 \\
HiDream-l1-Full \citep{cai2025hidream} & \checkmark & 0.543 & 0.024 & 0.284 \\
BLIP3-o \citep{chen2025blip3} & \checkmark & 0.021 & 0.018 & 0.020 \\
Janus-Pro \citep{chen2025janus} & \checkmark & 0.019 & 0.006 & 0.013 \\
\midrule
\textbf{ERNIE-Image-Turbo (w/o PE)} & \checkmark & 0.960 & 0.968 & 0.964 \\
\textbf{ERNIE-Image-Turbo (w/ PE)} & \checkmark & 0.968 & 0.964 & 0.966 \\
\textbf{ERNIE-Image (w/o PE)} & \checkmark & 0.968 & 0.959 & 0.964 \\
\textbf{ERNIE-Image (w/ PE)} & \checkmark & 0.980 & 0.966 & \underline{0.973} \\
\bottomrule
\end{tabular}
\end{table}

\begin{table}[H]
\centering
\caption{Quantitative comparison on the OneIG-EN benchmark \citep{chang2506oneig}. Best results are \textbf{bolded} and second-best results are \underline{underlined}.}
\label{tab:oneig_en}
\resizebox{\textwidth}{!}{
\begin{tabular}{lccccccc}
\toprule
Model & \makecell{Open\\Weights} & Alignment & Text & Reasoning & Style & Diversity & \makecell{Overall\\Score $\uparrow$} \\
\cmidrule(lr){3-7} \cmidrule(lr){8-8}
Nano Banana 2.0 \citep{nanobanana2026} & -- & 0.888 & 0.944 & 0.334 & \textbf{0.481} & 0.245 & \textbf{0.578} \\
Seedream 4.5 \citep{seedream4.5} & -- & \underline{0.891} & \textbf{0.998} & 0.350 & 0.434 & 0.207 & \underline{0.576} \\
Seedream 4.0 \citep{seedream2025seedream} & -- & \textbf{0.892} & 0.983 & 0.347 & 0.453 & 0.191 & 0.573 \\
Z-Image \citep{cai2025z} & \checkmark & 0.881 & 0.987 & 0.280 & 0.387 & 0.194 & 0.546 \\
Qwen-Image \citep{qwen-image} & \checkmark & 0.882 & 0.891 & 0.306 & 0.418 & 0.197 & 0.539 \\
GPT Image 1 [High] \citep{Gpt-image-1} & -- & 0.851 & 0.857 & 0.345 & \underline{0.462} & 0.151 & 0.533 \\
FLUX.2-klein-9B \citep{flux-2-2025} & \checkmark & 0.887 & 0.866 & 0.312 & 0.442 & 0.156 & 0.532 \\
Seedream 3.0 \citep{gao2025seedream} & -- & 0.818 & 0.865 & 0.275 & 0.413 & 0.277 & 0.530 \\
Qwen-Image-2512 \citep{qwen-image} & \checkmark & 0.876 & 0.990 & 0.292 & 0.338 & 0.151 & 0.530 \\
GLM-Image \citep{glm-image} & \checkmark & 0.805 & 0.969 & 0.298 & 0.353 & 0.213 & 0.528 \\
Z-Image-Turbo \citep{cai2025z} & \checkmark & 0.840 & \underline{0.994} & 0.298 & 0.368 & 0.139 & 0.528 \\
HiDream-l1-Full \citep{cai2025hidream} & \checkmark & 0.829 & 0.707 & 0.317 & 0.347 & 0.186 & 0.477 \\
Kolors 2.0 \citep{Kolors2.0} & \checkmark & 0.820 & 0.427 & 0.262 & 0.360 & \underline{0.300} & 0.434 \\
BAGEL \citep{bagel} & \checkmark & 0.769 & 0.244 & 0.173 & 0.367 & 0.251 & 0.361 \\
BLIP3-o \citep{chen2025blip3} & \checkmark & 0.711 & 0.133 & 0.223 & 0.361 & 0.229 & 0.307 \\
Janus-Pro \citep{chen2025janus} & \checkmark & 0.553 & 0.001 & 0.139 & 0.276 & \textbf{0.365} & 0.267 \\
\midrule
\textbf{ERNIE-Image-Turbo (w/o PE)} & \checkmark & 0.880 & 0.949 & 0.291 & 0.428 & 0.123 & 0.534 \\
\textbf{ERNIE-Image-Turbo (w/ PE)} & \checkmark & 0.868 & 0.967 & \underline{0.354} & 0.419 & 0.221 & 0.566 \\
\textbf{ERNIE-Image (w/o PE)} & \checkmark & \underline{0.891} & 0.967 & 0.295 & 0.447 & 0.169 & 0.554 \\
\textbf{ERNIE-Image (w/ PE)} & \checkmark & 0.868 & 0.979 & \textbf{0.357} & 0.431 & 0.241 & 0.575 \\
\bottomrule
\end{tabular}
}
\end{table}

\begin{table}[H]
\centering
\caption{Quantitative comparison on the OneIG-ZH benchmark \citep{chang2506oneig}. Best results are \textbf{bolded} and second-best results are \underline{underlined}.}
\label{tab:oneig_zh}
\resizebox{\textwidth}{!}{
\begin{tabular}{lccccccc}
\toprule
Model & \makecell{Open\\Weights} & Alignment & Text & Reasoning & Style & Diversity & \makecell{Overall\\Score $\uparrow$} \\
\cmidrule(lr){3-7} \cmidrule(lr){8-8}
Nano Banana 2.0 \citep{nanobanana2026} & -- & \textbf{0.843} & 0.983 & \textbf{0.311} & \textbf{0.461} & 0.236 & \textbf{0.567} \\
Seedream 4.0 \citep{seedream2025seedream} & -- & 0.836 & \underline{0.986} & 0.304 & 0.443 & 0.200 & \underline{0.554} \\
Seedream 4.5 \citep{seedream4.5} & -- & 0.832 & \underline{0.986} & 0.300 & 0.426 & 0.213 & 0.551 \\
Qwen-Image \citep{qwen-image} & \checkmark & 0.825 & 0.963 & 0.267 & 0.405 & 0.279 & 0.548 \\
Z-Image \citep{cai2025z} & \checkmark & 0.793 & \textbf{0.988} & 0.266 & 0.386 & 0.243 & 0.535 \\
Seedream 3.0 \citep{gao2025seedream} & -- & 0.793 & 0.928 & 0.281 & 0.397 & 0.243 & 0.528 \\
Qwen-Image-2512 \citep{qwen-image} & \checkmark & 0.823 & 0.983 & 0.272 & 0.342 & 0.157 & 0.515 \\
GLM-Image \citep{glm-image} & \checkmark & 0.738 & 0.976 & 0.284 & 0.335 & 0.221 & 0.511 \\
Z-Image-Turbo \citep{cai2025z} & \checkmark & 0.782 & 0.982 & 0.276 & 0.361 & 0.134 & 0.507 \\
GPT Image 1 [High] \citep{Gpt-image-1} & -- & 0.812 & 0.650 & 0.300 & \underline{0.449} & 0.159 & 0.474 \\
FLUX.2-klein-9B \citep{flux-2-2025} & \checkmark & 0.820 & 0.492 & 0.260 & 0.417 & 0.163 & 0.430 \\
Kolors 2.0 \citep{Kolors2.0} & \checkmark & 0.738 & 0.502 & 0.226 & 0.331 & \underline{0.333} & 0.426 \\
BAGEL \citep{bagel} & \checkmark & 0.672 & 0.365 & 0.186 & 0.357 & 0.268 & 0.370 \\
HiDream-l1-Full \citep{cai2025hidream} & \checkmark & 0.620 & 0.205 & 0.256 & 0.304 & 0.300 & 0.337 \\
BLIP3-o \citep{chen2025blip3} & \checkmark & 0.608 & 0.092 & 0.213 & 0.369 & 0.233 & 0.303 \\
Janus-Pro \citep{chen2025janus} & \checkmark & 0.324 & 0.148 & 0.104 & 0.264 & \textbf{0.358} & 0.240 \\
\midrule
\textbf{ERNIE-Image-Turbo (w/o PE)} & \checkmark & 0.833 & 0.909 & 0.258 & 0.400 & 0.132 & 0.506 \\
\textbf{ERNIE-Image-Turbo (w/ PE)} & \checkmark & 0.826 & 0.939 & 0.304 & 0.421 & 0.228 & 0.544 \\
\textbf{ERNIE-Image (w/o PE)} & \checkmark & \underline{0.842} & 0.898 & 0.266 & 0.421 & 0.177 & 0.521 \\
\textbf{ERNIE-Image (w/ PE)} & \checkmark & 0.830 & 0.954 & \underline{0.306} & 0.434 & 0.248 & \underline{0.554} \\
\bottomrule
\end{tabular}
}
\end{table}

\clearpage
\begin{figure}[t]
    \centering
    \includegraphics[width=0.73\linewidth]{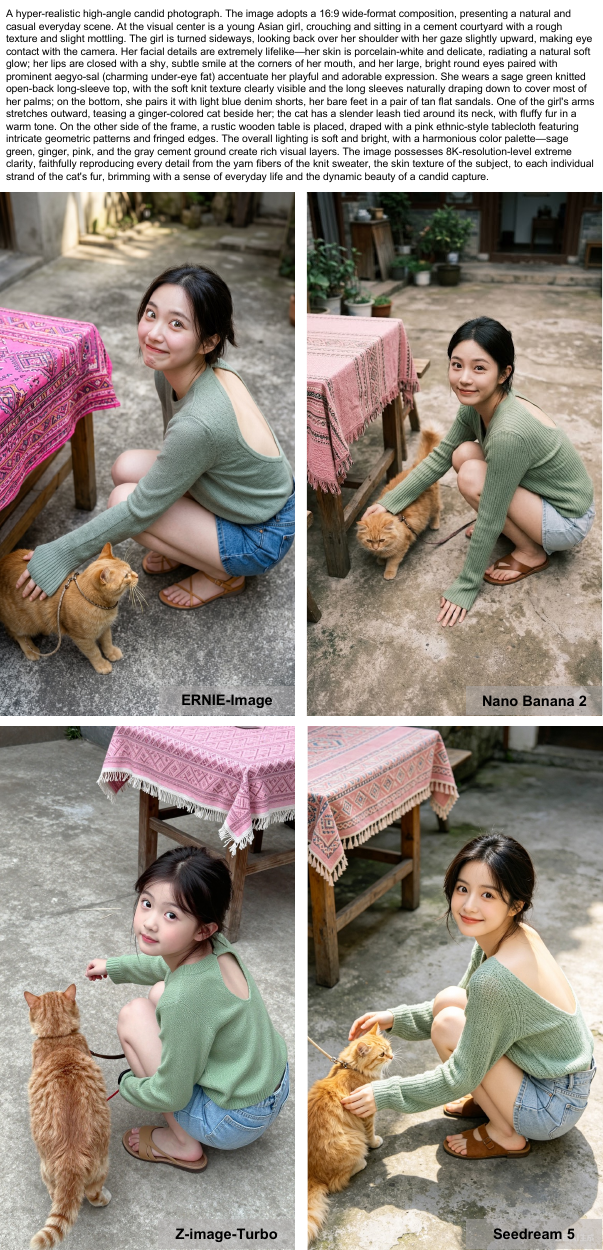}
    \caption{Comparison between ERNIE-Image and state-of-the-art open-source and closed-source models.}
    \label{fig:compare_photo1}
\end{figure}
\clearpage
\begin{figure}[t]
    \centering
    \includegraphics[width=0.8\linewidth]{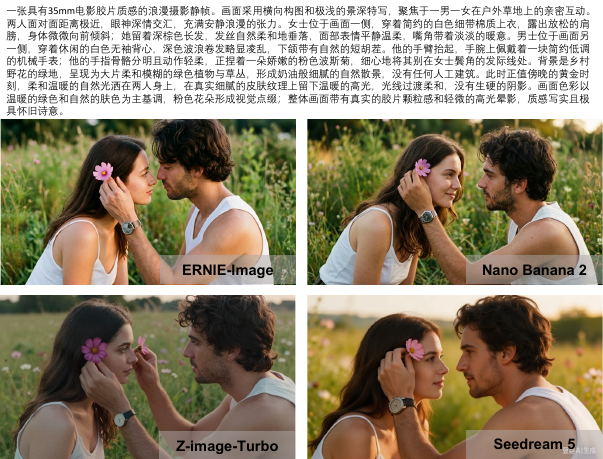}
    
    \vspace{0.5em}
    
    \includegraphics[width=0.8\linewidth]{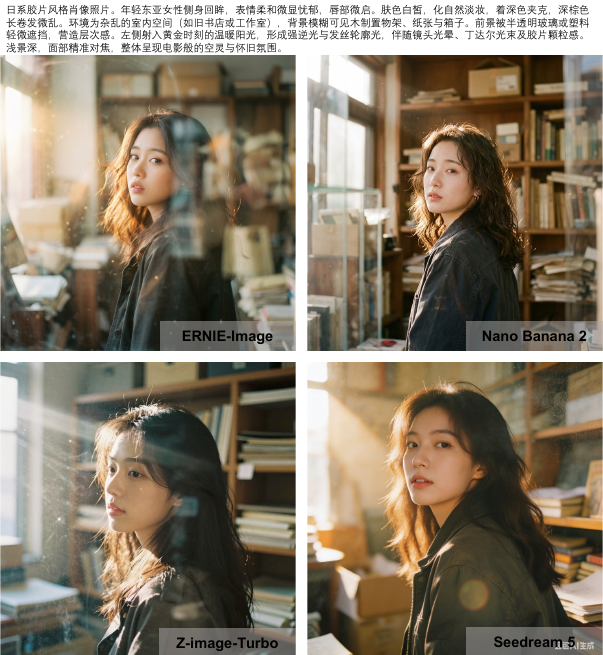}
    \caption{Comparison between ERNIE-Image and state-of-the-art open-source and closed-source models.}
    \label{fig:compare_photo2}
\end{figure}
\clearpage
\begin{figure}[t]
    \centering
    \includegraphics[width=0.8\linewidth]{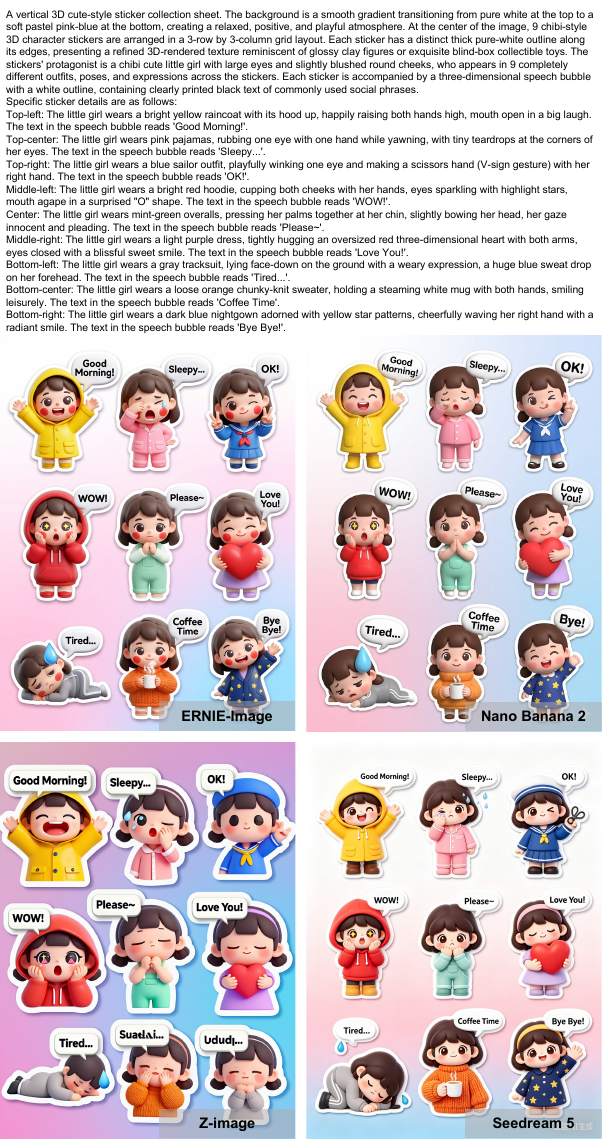}
    \caption{Comparison between ERNIE-Image and state-of-the-art open-source and closed-source models.}
    \label{fig:compare_instruction}
\end{figure}
\clearpage
\begin{figure}[t]
    \centering
    \includegraphics[width=0.7\linewidth]{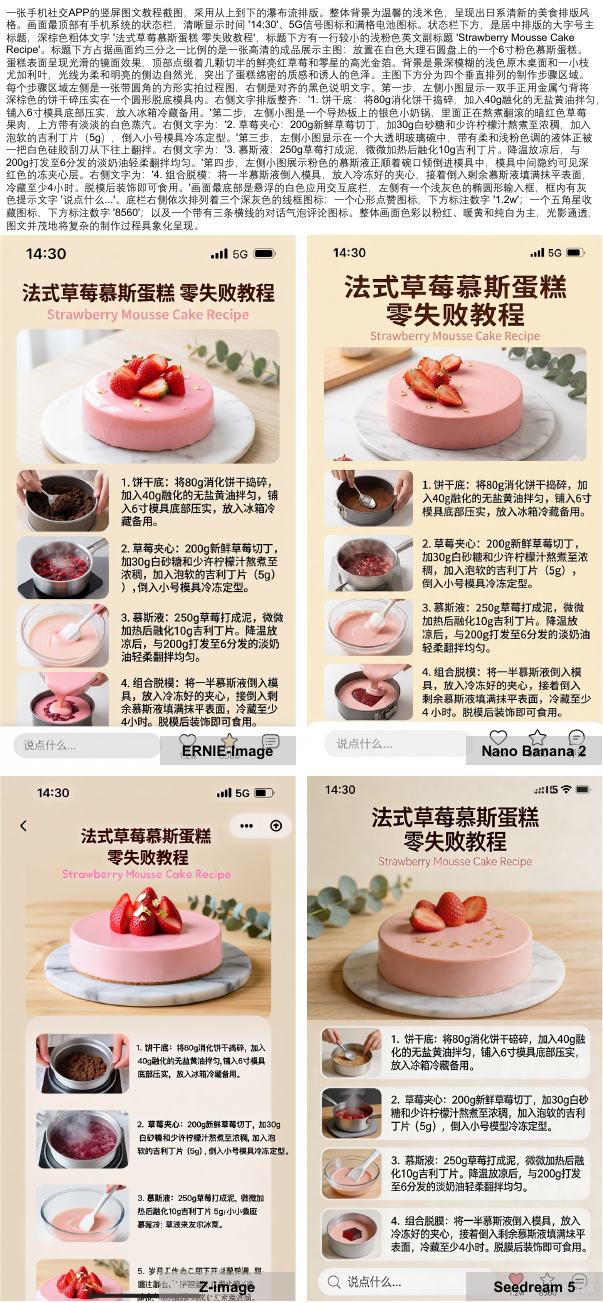}
    \caption{Comparison between ERNIE-Image and state-of-the-art open-source and closed-source models.}
    \label{fig:compare_text3}
\end{figure}
\clearpage
\begin{figure}[t]
    \centering
    \includegraphics[width=0.8\linewidth]{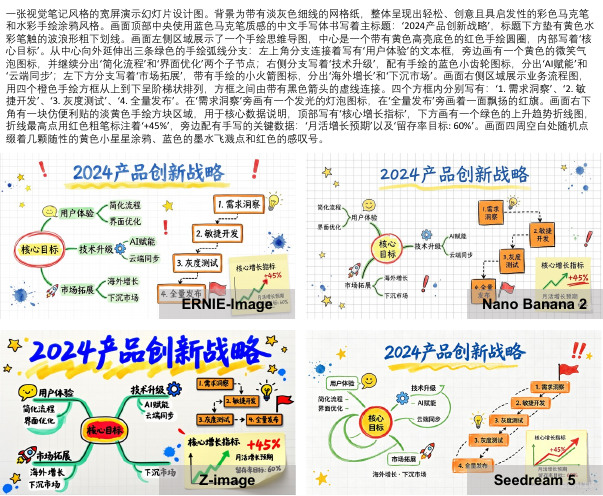}
    
    \vspace{0.5em}
    
    \includegraphics[width=0.8\linewidth]{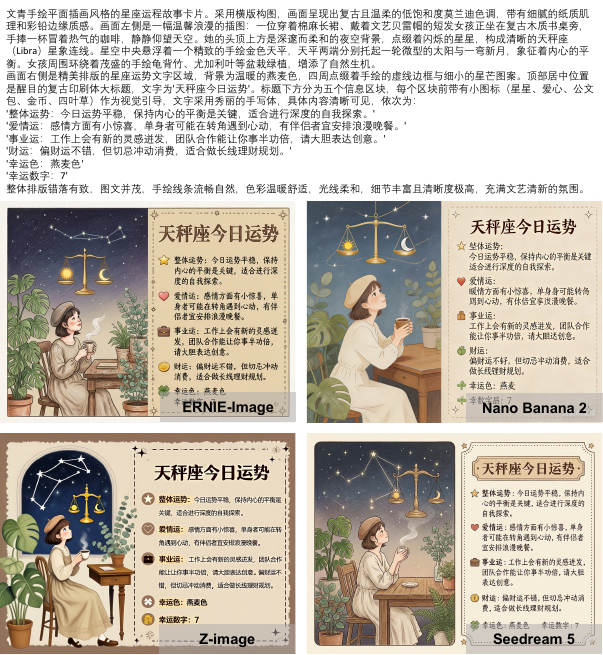}
    
    \caption{Comparison between ERNIE-Image and state-of-the-art open-source and closed-source models.}
    \label{fig:compare_text}
\end{figure}
\clearpage
\subsection{Qualitative Analysis}
To further demonstrate the capabilities of ERNIE-Image, we present a series of comparative examples against both open-source and proprietary models, including Z-Image series~\citep{cai2025z}, Seedream 5.0~\citep{seedream5}, and Nano Banana 2.0~\citep{nanobanana2026}. The evaluation focuses on three key aspects: portrait photography, complex instruction following, and text rendering.

\noindent \textbf{Portrait Photography.} As shown in Figure \ref{fig:compare_photo1}, ERNIE-Image produces highly realistic and lifelike images, accurately capturing the ``playful and adorable expression" specified in the prompt. Although Nano Banana 2.0 and Seedream 5.0 also generate semantically consistent results, they lack fine-grained facial details such as dimples, skin texture, and chin folds, resulting in images that appear polished yet less natural. In comparison, Z-Image-Turbo produces inferior results, with noticeable weaknesses in pose quality and image clarity.

\noindent \textbf{Complex Instruction Following.} The ability to follow complex instructions is critical for real-world design applications requiring fine-grained control. As illustrated in Figure \ref{fig:compare_instruction}, the prompt requires generating nine character stickers, each with distinct poses, outfits, and associated text. Under clear instructions, ERNIE-Image and Nano Banana produce largely similar outputs; however, Nano Banana incorrectly renders ``bye bye!" as ``bye!" for the final character. Seedream 5.0 misinterprets “scissors hand” as “holding scissors” in the third character, leading to semantic errors. Z-Image largely fails to follow the complex instructions and produces incorrect text. In contrast, ERNIE-Image accurately generates all characters with correct details and text, achieving a level of quality suitable for practical deployment.

\noindent \textbf{Chinese Text Rendering.} Due to the structural complexity and diversity of Chinese characters, text rendering remains a major challenge in text-to-image generation. In our evaluation, ERNIE-Image demonstrates robust performance (see Figure \ref{fig:compare_text3} and Figure \ref{fig:compare_text}), accurately rendering the specified Chinese text. Other models, however, often produce artifacts or incorrect characters, requiring multiple attempts and increasing usage cost.


\section{Conclusion}
In this report, we present the \textbf{ERNIE-Image} series, including a text-to-image diffusion model, a Prompt Enhancer, and an industrial-grade aesthetic model. We challenge the scaling law of model size and show that further mining pre-training data can push beyond the limits of conventional understanding. By introducing more complex and detailed instructions at an early stage of training, our model naturally emerges precise instruction-following and strong text-rendering capabilities, effectively surpassing many open-source and commercial models with far more extreme parameter sizes and costs. In terms of aesthetics, \textbf{ERNIE-Image-Aes} reduces the biases introduced by open-source datasets and annotations from specific groups, and can more faithfully reflect aesthetic quality that aligns with general public preferences, enabling ERNIE-Image to generate realistic photographs with diverse styles. With these improvements across multiple capabilities, our model achieves leading performance among open-source models on multiple benchmarks and human evaluations. Finally, we release all of the models we trained to promote open academic research and benefit the AIGC community, with the ultimate vision of making new technology easily accessible to everyone.

\section{Authors}

\textbf{Core Contributors}: 
\begin{itemize}
\item Lead: Jiaxiang Liu
\item Data: Zhida Feng, Pengyu Zou, Zhenyu Qian, Jiaxiang Liu, Yuehu Dong, Jun Xia, Yanzheng Lin, Honglin Xiong
\item Pre-training: Zhida Feng
\item Post-training: Zhenyu Qian, Zhida Feng, Pengyu Zou, Tianrui Zhu
\item Evaluation \& Analysis: Pengyu Zou
\item Deployment: Jun Xia, Yuehu Dong
\item Aesthetic Research: Jiaxiang Liu, Pengyu Zou, Zhenyu Qian

\end{itemize}

\textbf{Contributors}\textsuperscript{1}: 
Anqi Chen, Yunpeng Ding, Jinghui Duan, Lin Gao, Chao Han, Tiechao He, Jiakang Hu, Ranjun Hua, Xueming Jiang, Qingli Kong, Yuting Lei, Tianyu Li, Yunlin Liu, Changling Liu, Yaxin Liu, Yi Liu, Xuguang Liu, Xiaolong Ma, Yan Pan, Yiran Ren, Nan Sheng, Yu Sun, Siyang Sun, Yixiang Tu, Yang Wan, Huanai Wang, Siqi Wang, Yang Wu, Youzhi Yang, Xiaowen Yang, Jianwen Yang, Yehua Yang, Quanwen Zhang, Xinmin Zhang, Haoxin Zhang, Xiang Zhang, Jun Zhang, Qian Zhang, Qiao Zhao, Qi Zhou

\footnotetext[1]{Contributors are listed in alphabetical order of the last name.}

\clearpage %
\phantomsection %
\addcontentsline{toc}{section}{References} %
\bibliographystyle{colm2024_conference}
\bibliography{biblio}

\end{document}